\documentclass[10pt,twocolumn,letterpaper]{article}

\usepackage{wacv}

\makeatletter
\@namedef{ver@everyshi.sty}{}
\makeatother

\usepackage{xspace}
\usepackage{times}
\usepackage{epsfig}
\usepackage{graphicx}
\usepackage{amsmath, amssymb, amsthm}
\usepackage{booktabs}
\usepackage{array}
\usepackage{multirow}
\usepackage[flushleft]{threeparttable}
\usepackage[table,usenames,dvipsnames]{xcolor}
\usepackage[accsupp]{axessibility}
\usepackage{soul}
\sethlcolor{lightgray}
\usepackage[title]{appendix}

\makeatletter
\def\input@path{{src/}{./}}
\makeatother

\graphicspath{{./fig/}}

\newcommand{\paren}[1]{\left(#1\right)}

\newcommand{\acl}{{ACL-GAN}\xspace}
\newcommand{\cou}{{Council-GAN}\xspace}
\newcommand{\cyc}{{CycleGAN}\xspace}
\newcommand{\uga}{{U-GAT-IT}\xspace}
\newcommand{\vitgan}{{UVCGAN}\xspace} 
\newcommand{\uv}{UNet-ViT\xspace}

\newcommand{\anime}{\textrm{Selfie2Anime}\xspace}
\newcommand{\gender}{\textrm{GenderSwap}\xspace}
\newcommand{\glasses}{\textrm{Eyeglasses}\xspace}

\newcommand{\repo}{\url{https://github.com/LS4GAN/uvcgan}}

\usepackage{tikz}
\usepackage{pgfplots}
\pgfplotsset{compat=1.17} 
\usetikzlibrary{
    fit,
    calc,
    math,
    patterns,
    external,
    backgrounds,
    arrows.meta,
    shadows.blur,
}
\definecolor{cooldark}{HTML}{370665}
\definecolor{coollight}{HTML}{35589A}
\definecolor{warmdark}{HTML}{F14A16}
\definecolor{warmlight}{HTML}{FC9918}
\definecolor{other}{HTML}{509B7D} 

\pgfdeclarelayer{back}
\pgfsetlayers{back,main}
\def\wacvPaperID{184}
\wacvalgorithmstrack
\wacvfinalcopy
\ifwacvfinal
\usepackage[breaklinks=true,bookmarks=false]{hyperref}
\else
\usepackage[pagebackref=true,breaklinks=true,colorlinks,bookmarks=false]{hyperref}
\fi

\begin{document}

\title{UVCGAN: UNet Vision Transformer cycle-consistent GAN for unpaired image-to-image translation}
\author{
    Dmitrii Torbunov, Yi Huang, Haiwang Yu, Jin Huang, \\ Shinjae Yoo, Meifeng Lin, Brett Viren, Yihui Ren\\
    Brookhaven National Laboratory, Upton, NY, USA\\
    {\tt\small dtorbunov,yhuang2,hyu,jhuang,sjyoo,mlin,bviren,yren@bnl.gov}
}


\maketitle

\begin{abstract}

    Unpaired image-to-image translation has broad applications in art, design, and scientific simulations. 
    One early breakthrough was \cyc that emphasizes one-to-one mappings between two unpaired image domains via generative-adversarial networks (GAN) coupled with the cycle-consistency constraint, 
    while more recent works promote one-to-many mapping to boost diversity of the translated images. 
    Motivated by scientific simulation and one-to-one needs, this work revisits the classic \cyc framework and boosts its performance to outperform more contemporary models \emph{without relaxing the cycle-consistency constraint}. 
    To achieve this, we equip the generator with a Vision Transformer (ViT) and employ necessary training and regularization techniques. 
    Compared to previous best-performing models, our model performs better and retains a strong correlation between the original and translated image. 
    An accompanying ablation study shows that both the gradient penalty and self-supervised pre-training are crucial to the improvement.
    To promote reproducibility and open science, the source code, hyperparameter configurations, and pre-trained model are available at 
    \repo.

\end{abstract}


\section{Introduction}
\label{sec:intro}

Deep generative models such as generative adversarial networks
(GAN)~\cite{NIPS2014_5ca3e9b1,brock_large_2019,karras2020analyzing},
variational autoencoder (VAE)~\cite{kingma_auto-encoding_2013,kingma_introduction_2019},
normalizing flow (NF)~\cite{dinh_density_2017,kingma_glow_2018}, and
diffusion models (DM)~\cite{ho_denoising_2020,song_denoising_2021,nichol_glide_2022} represent a
class of statistical models used to create realistic and diverse data instances that mimic ones from a target data domain. 
Along with applications in image processing, audio analysis, and text generation, 
their success and expressiveness have attracted researchers in natural science, including cosmology~\cite{mustafa_cosmogan_2019}, high-energy physics~\cite{di_sipio_dijetgan_2019,alanazi_simulation_2021}, 
materials design~\cite{fuhr_deep_2022}, and 
drug design~\cite{das_accelerated_2021,bilodeau_generative_nodate}.
Most existing work treats deep generative models as drop-in replacements for existing
simulation software. 
Modern simulation frameworks can generate data with high fidelity, yet the data are imperfect. 
Widespread systematic inconsistencies between the generated and actual data significantly limit the applicability of simulation results. 
We would like to take advantage of the expressiveness of deep generative models to bridge this simulation versus reality gap.
We frame the task as an unpaired
image-to-image translation problem, where simulation results
can be defined as one domain with experimental data as
the other. 
Unpaired is a necessary constraint because gathering simulation and experiment data with exact pixel-to-pixel mapping is difficult (often impossible).
Apart from improving the quality of the simulation results, the successful generative model can be run in the inverse direction to translate real-world data into the simulation domain. This inverse task can be viewed as a denoising step, helpful toward correctly inferring the underlying parameters from experiment observations~\cite{cranmer_frontier_2020}.
Achieving realistic scientific simulations requires both well-defined scientific datasets and purposefully designed machine learning models. 
This work will focus on the latter by developing novel models for unpaired image-to-image translation.

The \cyc~\cite{zhu_unpaired_2017} model is the first of its kind to 
translate images between two domains without paired instances. 
It uses two GANs, one for each translation direction. 
\cyc introduces cycle-consistency loss, where an image should look
like itself after a cycle of translations to the other domain and back.
Such cycle-consistency is of utmost importance for scientific applications as the science cannot be altered during translation.
Namely, there should be a one-to-one mapping between a simulation result and its experimental counterpart.
However, to promote more diverse image generation, many
recent works~\cite{zhao2020unpaired,nizan_breaking_2019,park_contrastive_2020,zhao2021unpaired} 
relaxed the cycle-consistency constraint. 
Following the same objective of revisiting and modifying canonical neural architectures~\cite{bello2021revisiting},
we demonstrate that
by equipping \cyc with a Vision Transformer (ViT)~\cite{dosovitskiy2020image} to boost non-local pattern learning
and employing advanced training techniques, such as gradient penalty and self-supervised pre-training,
the resulting model, named UVCGAN, can outperform competing models in several benchmark datasets.

\noindent \textbf{Contributions.}
In this work, we: 1) incorporated ViT to the CycleGAN generator and employed advanced training techniques, 2) demonstrated its superb image translation performance versus other more 
heavy models, 3) showed via an ablation study that the architecture change alone is insufficient to compete with other methods and pre-training and gradient-penalty are needed, and 4) identified the unmatched evaluation results from past literature and standardized the evaluation procedure to ensure a fair comparison and promote reusability of our benchmarking results.

\section{Related work}
\label{sec:related work}



\noindent\textbf{Deep Generative Models.} 
Deep generative models create realistic data points (images, molecules, audio samples,
etc.) that are similar to those presented in a dataset. 
Unlike decision-making models that contract representation dimension and distill high-level information,
generative models enlarge the representation dimension and extrapolate information.
There are several types of deep generative models.
A VAE~\cite{kingma_auto-encoding_2013,kingma_introduction_2019,liu_constrained_2018,razavi_generating_2019} reduces data points into a probabilistic latent space
and reconstructs them from samples of latent distributions. 
NFs~\cite{dinh_density_2017,kingma_glow_2018,chen_residual_2019,grathwohl_ffjord_2022} make use of the change of variable formula and transform samples from a normal distribution 
to the data distribution via a sequence of invertible and differentiable transformations. 
DMs~\cite{ho_denoising_2020,song_denoising_2021,nichol_glide_2022,rombach_high-resolution_2022,salimans_progressive_2022,watson_learning_2022} are parameterized Markov chains trained to transform noise 
into data (forward process) via successive steps. 
Meanwhile, GANs~\cite{NIPS2014_5ca3e9b1} formulate the learning process as a minimax game, where the generator tries to fool the discriminator by creating realistic data points, and the discriminator attempts to distinguish the generated samples from the real ones. 
GANs are among the most expressive and flexible models that can generate high-resolution, diverse, style-specific images~\cite{brock_large_2019,karras2020analyzing}.

\noindent\textbf{GAN Training Techniques.} %
The original GAN suffered from many problems, such as mode collapsing and training divergence~\cite{mescheder2018training}.
Since then, much work has been done to improve training stability and model diversity.
ProGAN~\cite{karras_progressive_2018} introduces two stabilization methods: progressive training and learning rate equalization.
Progressive training of the generator starts
from low-resolution images and moves up to high-resolution ones.
The learning rate equalization scheme seeks to ensure that
all parts of the model are being trained at the same rate.
Wasserstein GAN~\cite{gulrajani2017improved} suggests that the destructive
competition between the generator and discriminator can be prevented
by using a better loss function, i.e., the Wasserstein loss function. Its key ingredient is a gradient penalty term that
prevents the magnitude of the discriminator gradients from growing too large. 
However, the Wasserstein loss function was later reexamined. Notably, the assessment revealed the gradient penalty term was responsible for stabilizing the training and not the Wasserstien loss function~\cite{stanczuk2021wasserstein}.
In addition, the StyleGAN~v2~\cite{karras2020analyzing} relies on a
zero-centered gradient penalty term to achieve state-of-the-art results on a
high-resolution image generation task. 
These findings motivated this work  to explore applying the gradient penalty terms to improve GAN training stability.


\noindent\textbf{Transformer Architecture for Computer Vision.}
Convolutional neural network (CNN) architecture is a popular choice for computer vision tasks. 
In the natural language processing (NLP) field, the attention mechanism and transformer-style architecture have surpassed previous models, such as hidden Markov models and recurrent neural networks, in open benchmark tasks. 
Compared to CNNs, transformers can more efficiently capture non-local patterns, which are common in nature. 
Applications of transformers in computer vision debuted in~\cite{dosovitskiy2020image}, while other recent work has shown that a CNN-transformer hybrid can achieve better performance~\cite{xiao_early_2021,guo_cmt_2021}.

\noindent\textbf{Self-supervised Pre-training.} 
Self-supervised pre-training primes network initial weights by training 
the network on artificial tasks derived from the original data without supervision.
This is especially important for training models with a large amount of parameters on 
a small labeled dataset as they tend to overfit. 
There are many innovative ways to create these artificial self-supervision tasks.
Examples in computer vision include image inpainting~\cite{pathak_context_2016}, solving jigsaw puzzles~\cite{noroozi_unsupervised_2016}, 
predicting image rotations~\cite{komodakis_unsupervised_2018}, 
multitask learning~\cite{doersch_multi-task_2017},  
contrastive learning~\cite{chen_simple_2020,chen_exploring_2021}, 
and teacher-student latent bootstrapping~\cite{grill2020bootstrap,caron2021emerging}.
Common pre-training methods in NLP include
the auto-regressive~\cite{radford2018improving} and mask-filling~\cite{devlin2019bert} tasks.
In the mask-filling task, some parts of the sentence are masked, and the network is tasked
with predicting the missing parts from their context. 
Once a model is pre-trained, it can be fine-tuned for multiple downstream tasks using much smaller labeled datasets.


We hypothesise GAN training also can benefit from self-supervised pre-training.
In particular, GAN training is known to suffer from the ``mode collapse'' problem~\cite{mescheder2018training}: 
the generator fails to reproduce the target distribution of images faithfully. Instead, only a small set of images are generated repeatedly despite diverse input samples. 
Observations have noted the mode collapse
problem occurs just a few epochs after beginning the GAN training~\cite{karras_progressive_2018}.
This suggests that better initialized model weights could be used.
Indeed, transfer learning of GANs, a form of pre-training, has been an effective way to improve GAN performance on small training datasets~\cite{wang_transferring_2018,noguchi_image_2019,zhao_leveraging_2020,wang_minegan_2020,grigoryev_when_2022}.
However, scientific data, such as those in cosmology and high energy physics, are remotely similar to natural images.
Therefore, we have chosen only to pre-train generators on a self-supervised inpainting task, which has been successful in both NLP and computer vision.
Moreover, it is well suited for image-to-image translation models, where the model's output shape is the same as its input shape.

\noindent\textbf{GAN Models for Unpaired Image-to-image Translation.}
Many frameworks~\cite{huang_multimodal_2018,lee_diverse_2018,nizan_breaking_2019,zhao2020unpaired} have been developed for unpaired image-to-image translation.
While most commonly use GANs for translation,
they differ in how consistency is maintained. 
\uga~\cite{kim_u-gat-it_2020} follows the \cyc closely but relies on more sophisticated generator and discriminator networks for better performance. 
Other models relax the cycle-consistency constraint.
For example, ACL-GAN~\cite{zhao2020unpaired} relaxes the per-image consistency constraint by introducing the so-called ``adversarial-consistency loss'' that imposes cycle-consistency at a distribution level between a neighborhood of the input and the translations. Meanwhile, \cou~\cite{nizan_breaking_2019} 
abandons the idea of explicit consistency enforcement and instead relies on a generator ensemble with the assumption that, when multiple generators arrive at an agreement, the commonly agreed upon portion is what should be kept consistent.
While relaxed or implicit consistency constraints boost translation diversity and achieve better evaluation scores, such models inevitably introduce  randomness into the feature space and output. Hence, they are unsuitable for applications where a one-to-one mapping is required. 
Compared to the original CycleGAN, 
all these models contain more parameters requiring more computation resources and longer training time.
Concurrently, Zheng et. al.~\cite{zheng2022ittr} also proposed to utilize ViT for image translation by replacing the ResNet blocks with hybrid blocks of self-attention and convolution. 

\section{Method}
\subsection{\cyc-like Models}
\label{sec:cyc-like_models}

\begin{figure}[ht]
    \centering
    \tikzsetnextfilename{cyclegan}
    \vspace{-5pt}
\resizebox{\linewidth}{!}{
\begin{tikzpicture}[font={\fontsize{14}{14}\selectfont}]
    
    \tikzstyle{block} = [
        rectangle, 
        rounded corners=3pt, 
        draw=white,
        fill=white,
        blur shadow={shadow blur steps=5, shadow xshift=1pt, shadow yshift=-1pt}
    ]
    
    \def\xs{24pt}
    \def\ys{54pt}
    \tikzstyle{tb} = [block,align=center]
    \tikzstyle{input} = [tb,circle,inner sep=2pt, minimum size=26pt,draw=black!30,fill=white]
    \tikzstyle{output} = [tb,circle,inner sep=2pt, minimum size=26pt,draw=black!30,fill=black!20]
    \tikzstyle{generator} = [tb,inner sep=5pt,text=white,draw=other,fill=other!80]
    \tikzstyle{discriminator} = [tb,inner sep=5pt,text=white,draw=coollight,fill=coollight!80,anchor=center]
    \tikzstyle{loss} = [tb,draw=none,fill=black!20,inner sep=5pt,align=left,anchor=center,execute at begin node=\setlength{\baselineskip}{3ex}]
    \tikzstyle{flow} = [-{Latex[scale=.6]},rounded corners=10pt, line width=2pt]

    \node[input] (A) at (0, 0) {$A$};
    \node[generator,anchor=west] (GabU) at ([xshift=\xs]A.east) {$\mathcal{G}_{A\rightarrow B}$};
    \node[output,anchor=west] (FB) at ([xshift=\xs]GabU.east) {$B_f$};
    \node[generator,anchor=west] (GbaU) at ([xshift=\xs]FB.east) {$\mathcal{G}_{B\rightarrow A}$};
    \node[output,anchor=west] (CA) at ([xshift=\xs]GbaU.east) {$A_c$};
    \draw[flow] (A) -- (GabU);
    \draw[flow] (GabU) -- (FB);
    \draw[flow] (FB) -- (GbaU);
    \draw[flow] (GbaU) -- (CA);
    \node[loss,anchor=south,fill=warmdark!40] (CycleLossU) at ([yshift=.35*\ys]FB.north) {cycle-consistency loss};
    \draw[flow] (A) |- (CycleLossU);
    \draw[flow] (CA) |- (CycleLossU);

    \node[input,anchor=north] (B) at ([yshift=-\ys]FB.south) {$B$};
    \node[generator,anchor=east] (GbaL) at ([xshift=-\xs]B.west) {$\mathcal{G}_{B\rightarrow A}$};
    \node[output,anchor=east] (FA) at ([xshift=-\xs]GbaL.west) {$A_f$};
    \node[generator,anchor=east] (GabL) at ([xshift=-\xs]FA.west) {$\mathcal{G}_{A\rightarrow B}$};
    \node[output,anchor=east] (CB) at ([xshift=-\xs]GabL.west) {$B_c$};
    \draw[flow] (B) -- (GbaL);
    \draw[flow] (GbaL) -- (FA);
    \draw[flow] (FA) -- (GabL);
    \draw[flow] (GabL) -- (CB);
    \node[loss,anchor=north,fill=warmdark!40] (CycleLossL) at ([yshift=-.35*\ys]FA.south) {cycle-consistency loss};
    \draw[flow] (B) |- (CycleLossL);
    \draw[flow] (CB) |- (CycleLossL);
    
    \node[discriminator] (DA) at ($(A)!.5!(FA)$) {$\mathcal{D}_A$};
    \node[discriminator] (DB) at ($(FB)!.5!(B)$) {$\mathcal{D}_B$};
    \draw[flow] (A) -- (DA);
    \draw[flow] (FA) -- (DA);
    \draw[flow] (B) -- (DB);
    \draw[flow] (FB) -- (DB);
    
    \node[input] (AA) at ([xshift=-7*\xs]DA.west) {$A$};
    \node[generator,anchor=west] (GbaI) at ([xshift=\xs]AA.east) {$\mathcal{G}_{B\rightarrow A}$};
    \node[output,anchor=west] (AI) at ([xshift=\xs]GbaI.east) {$A_i$};
    \draw[flow] (AA) -- (GbaI);
    \draw[flow] (GbaI) -- (AI);
    \node[loss,anchor=south,fill=warmdark!40] (AIdtLoss) at ([yshift=.35*\ys]GbaI.north) {identity loss};
    \draw[flow] (AA) |- (AIdtLoss);
    \draw[flow] (AI) |- (AIdtLoss);
    
    \node[input] (BB) at ([xshift=2*\xs]DB.east) {$B$};
    \node[generator,anchor=west] (GabI) at ([xshift=\xs]BB.east) {$\mathcal{G}_{B\rightarrow A}$};
    \node[output,anchor=west] (BI) at ([xshift=\xs]GabI.east) {$B_i$};
    \draw[flow] (BB) -- (GabI);
    \draw[flow] (GabI) -- (BI);
    \node[loss,anchor=north,fill=warmdark!40] (BIdtLoss) at ([yshift=-.35*\ys]GabI.south) {identity loss};
    \draw[flow] (BB) |- (BIdtLoss);
    \draw[flow] (BI) |- (BIdtLoss);
    
    \node[loss,inner ysep=5pt,inner xsep=10pt,anchor=south west,fill=warmdark!40] (LegendGLoss) at ([xshift=-\xs]BB |- CycleLossL.south) {};
    \node[inner sep=0, anchor=west] at ([xshift=5pt]LegendGLoss.east) {generator losses};
    \node[loss,inner ysep=5pt,inner xsep=10pt,anchor=south west,fill=warmlight!40] (LegendDLoss) at ([yshift=5pt]LegendGLoss.north west) {};
    \node[inner sep=0, anchor=west] at ([xshift=5pt]LegendDLoss.east) {discriminator  loss};

    \tikzmath{
        coordinate \C;
        \C = (BI.east |- CycleLossU.north)-(AA.west |- CycleLossL.south); 
    }
    \begin{pgfonlayer}{back}
        \node[rounded corners=5pt,inner xsep=.505*\Cx,inner ysep=.505*\Cy,draw=none,fill=white, anchor=center] (BG) at ($(DA)!.5!(DB)$) {};
    \end{pgfonlayer}
    
    \tikzmath{
        coordinate \C;
        \C = (A.north -| DA.east)-(FA.south -| DA.west); 
    }
    \begin{pgfonlayer}{back}
        \node[rounded corners=5pt,inner xsep=.96*\Cx,inner ysep=.61*\Cy,draw=none,fill=warmlight!40,anchor=center] (DALoss) at (DA) {};
        \node[rounded corners=5pt,inner xsep=.74*\Cx,inner ysep=.35*\Cy,draw=none,fill=warmdark!40,anchor=center] (GabGANLoss) at ($(A)!.5!(DA)$) {};
    \end{pgfonlayer}
    
    \tikzmath{
        coordinate \C;
        \C = (FB.north -| DA.east)-(B.south -| DA.west); 
    }
    \begin{pgfonlayer}{back}
        \node[rounded corners=5pt,inner xsep=.96*\Cx,inner ysep=.60*\Cy,draw=none,fill=warmlight!40,anchor=center] (DBLoss) at (DB) {};
        \node[rounded corners=5pt,inner xsep=.74*\Cx,inner ysep=.35*\Cy,draw=none,fill=warmdark!40,anchor=center] (GbaGANLoss) at ($(B)!.5!(DB)$) {};
    \end{pgfonlayer}
    
\end{tikzpicture} 
}

    \caption{\textbf{\cyc Framework}}
    \label{fig:cyclegan}
\end{figure}


\def\genab{\mathcal{G}_{A\rightarrow B}}
\def\genba{\mathcal{G}_{B\rightarrow A}}
\def\da{\mathcal{D}_A}
\def\db{\mathcal{D}_B}

\cyc-like models~\cite{zhu_unpaired_2017,kim_u-gat-it_2020}
interlace two generator-discriminator pairs for unpaired image-to-image translation (\autoref{fig:cyclegan}).
Denote the two image domains by $A$ and $B$, a \cyc-like model uses generator $\genab$ to translate images from $A$ to $B$, and generator $\genba$, $B$ to $A$. Discriminator $\da$ is used to distinguish between images in $A$ and those translated from $B$ (denoted as $A_f$ in~\autoref{fig:cyclegan}) and discriminator $\db$, $B$ and $B_f$.

The discriminators are updated by backpropagating loss corresponding to failure in distinguishing real and translated images (called \textit{generative adversarial loss} or \textit{GAN loss}): \
\vspace{-10pt}
\begin{align}
    \mathcal{L}_{\text{disc}, A} = & \mathbb{E}_{x\sim{B}}\ell_{\text{GAN}}\paren{\da\paren{\genba(x)}, 0}\nonumber\\ 
    &+ \mathbb{E}_{x\sim{A}}\ell_\text{GAN}\paren{\da(x), 1}, \label{eq:disc_loss_A}\\
    \mathcal{L}_{\textrm{disc}, B} = & \mathbb{E}_{x\sim{A}}\ell_{\text{GAN}}\paren{\db\paren{\genab(x)}, 0}\nonumber\\ 
    & + \mathbb{E}_{x\sim{B}}\ell_\text{GAN}\paren{\db(x), 1} \label{eq:disc_loss_B}.
\end{align}
Here, $\ell_{\text{GAN}}$ can be any classification loss function ($L_2$, cross-entropy, Wasserstein~\cite{arjovsky2017wasserstein}, etc.), while the $0$ and $1$ are class labels for translated (fake) and real images, respectively. 
The generators are updated by backpropagating loss from three sources: GAN loss, cycle-consistency loss, and identity-consistency loss. Using $\genab$ as an example:
\begin{align}
    \mathcal{L}_{\text{GAN}, A} =& \mathbb{E}_{x\sim{A}}\ell_\text{GAN}\paren{\da\paren{\genab(x)}, 1},\\
    \mathcal{L}_{\text{cyc}, A} =& \mathbb{E}_{x\sim{A}}\ell_\text{reg}\paren{\genba\paren{\genab(x)}, x},\\
    \mathcal{L}_{\text{idt}, A} =& \mathbb{E}_{x\sim{A}} \ell_\text{reg}\paren{\genba\paren{x}, x}.
\end{align}
And,
\begin{align}
    \mathcal{L}_{\text{gen}, A\rightarrow B} =& \mathcal{L}_{\text{GAN}, A} + \lambda_\text{cyc}\mathcal{L}_{\text{cyc}, A} + \lambda_\text{idt}\mathcal{L}_{\text{idt}, A}, \label{eq:gen_loss_A}\\ 
    \mathcal{L}_{\text{gen}, B\rightarrow A} =& \mathcal{L}_{\text{GAN}, B} + \lambda_\text{cyc}\mathcal{L}_{\text{cyc}, B} + \lambda_\text{idt}\mathcal{L}_{\text{idt}, B}. \label{eq:gen_loss_B}
\end{align}
Here, $\ell_\textrm{reg}$ can be any regression loss function ($L_1$ or $L_2$, etc.), and $\lambda_\text{cyc}$ and $\lambda_\text{idt}$ are combination coefficients. 

To improve the original \cyc model's performance, we implement three major changes.
First, we modify the generator to have a hybrid architecture based on a UNet
with a ViT bottleneck (Section~\ref{subsec:gen}). Second, to regularize the \cyc discriminator, we augment the vanilla \cyc discriminator loss with a gradient penalty term (Section~\ref{subsec:gradientPenalty}). 
Finally, instead of training from a randomly initialized network weights, we pre-train generators in a self-supervised fashion on the image inpainting task to obtain a better starting state (Section~\ref{subsec:pretraining}). 
\subsection{\uv Generator}
\label{subsec:gen}

\begin{figure*}[ht]
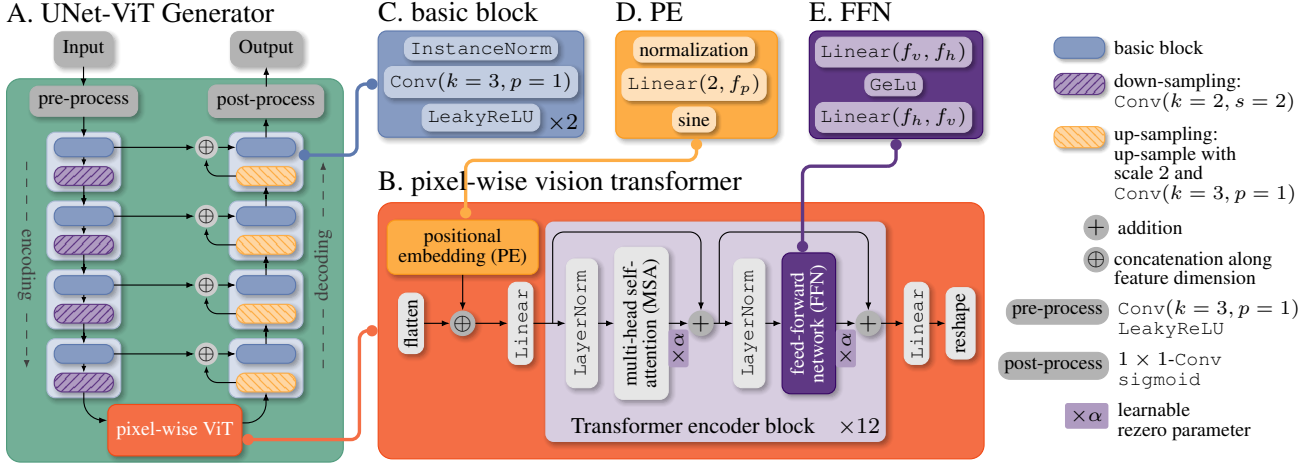

    \centering
    \tikzsetnextfilename{gen}
    
    \resizebox{\textwidth}{!}{
        \begin{tikzpicture}
            \node[inner sep=0, fill=white] at (0, 0) {


\begin{tikzpicture}

\tikzstyle{block} = [
    rectangle, 
    rounded corners=3pt, 
    draw=white,
    fill=white,
    blur shadow={shadow blur steps=5, shadow xshift=1pt, shadow yshift=-1pt}
]
\tikzstyle{tb} = [block,align=center,font=\fontsize{7}{7}\selectfont]
\tikzstyle{flow} = [-{Latex[scale=.8]}, rounded corners=2pt]
\def\tys{3pt}

\def\gcolor{other!80}
\node[inner sep=2pt,draw=other,fill=\gcolor,rounded corners=5pt] (U) at (0, 0) {\input{fig/unet_small}};

\tikzstyle{inout} = [block,font=\fontsize{7}{7}\selectfont,anchor=south, inner sep=3pt]
\def\xsio{32.5pt}
\node[inout,draw=black!30,fill=black!30] (I) at ([xshift=-\xsio]$(U.south)!1.03!(U.north)$) {Input};
\node[inout,draw=black!30,fill=black!30] (O) at ([xshift=\xsio]$(U.south)!1.03!(U.north)$) {Output};
\draw[-{Latex[scale=.6]}] (I) -- ([xshift=-\xsio,yshift=-3pt]U.north);
\draw[-{Latex[scale=.6]}] ([xshift=\xsio,yshift=-3pt]U.north) -- (O);

\node[inner sep=0,anchor=south west] (LU) at ([yshift=\tys]U.west |- I.north) {A.~UNet-ViT Generator};

\node[block,anchor=north west,inner sep=2pt,draw=coollight!80,fill=coollight!60] (B) at ([yshift=17pt]$(U.north west)!1.1!(U.north east)$) {
\begin{tikzpicture}
    \tikzstyle{basicblockdetail} = [tb,anchor=north,inner sep=2pt,draw=coollight!30,fill=coollight!30]
    \node[basicblockdetail] (Norm) at (0, 0) {\texttt{InstanceNorm}};
    \node[basicblockdetail] (Conv) at ([yshift=-2pt]Norm.south) {\texttt{Conv}$(k=3, p=1)$};
    \node[basicblockdetail] (Activ) at ([yshift=-2pt]Conv.south) {\texttt{LeakyReLU}};
\end{tikzpicture}
};
\node[inner sep=0, anchor=south west] (BTitle) at ([yshift=\tys]B.north west) {C.~basic block};
\node[inner sep=0, anchor=south east] at ([xshift=-2pt,yshift=2pt]B.south east) {\footnotesize$\times2$};
\draw[{Circle[coollight!80,length=4pt]}-{Circle[coollight!80,length=4pt]},coollight!80,very thick,rounded corners=3pt] ($(B.north west)!.5!(B.south west)$) -| ([yshift=-6pt]$(B.east)!1.1!(B.west)$) |- ([xshift=44pt]$(U.north)!.2!(U.south)$);

\node[tb,anchor=north west,inner sep=2pt,draw=warmlight,fill=warmlight!80] (PE) at ([xshift=11pt]B.north east) {
\begin{tikzpicture}
    \tikzstyle{PEdetail} = [tb,anchor=north,inner sep=2pt,draw=warmlight!30,fill=warmlight!30]
    \node[PEdetail] (norm) at (0, 0) {normalization};
    \node[PEdetail] (linear) at ([yshift=-2pt]norm.south) {\texttt{Linear}$(2, f_p)$};
    \node[PEdetail] (sine) at ([yshift=-2pt]linear.south) {sine};
\end{tikzpicture}
};
\node[inner sep=0, anchor=south west] at ([yshift=\tys]PE.north west) {D.~PE};

\node[tb,anchor=north west,inner sep=2pt,draw=cooldark,fill=cooldark!80] (FFN) at ([xshift=11pt]PE.north east) {
\begin{tikzpicture}
    \tikzstyle{Fdetail} = [tb,anchor=north,inner sep=1.8pt,draw=cooldark!30,fill=cooldark!30]
    \node[Fdetail] (FdL1) at (0, 0) {\texttt{Linear}$(f_v, f_h)$};
    \node[Fdetail] (FdA) at ([yshift=-2pt]FdL1.south) {\texttt{GeLu}};
    \node[Fdetail] (FdL2) at ([yshift=-2pt]FdA.south) {\texttt{Linear}$(f_h, f_v)$};
\end{tikzpicture}
};
\node[inner sep=0,anchor=south west] (FTitle) at ([yshift=\tys]FFN.north west) {E.~FFN};

\node[inner xsep=3pt,inner ysep=6pt,anchor=north west,draw=warmdark,fill=warmdark!80,rounded corners=5pt] (V) at ($(B.north west)!1.6!(B.south west)$) {\input{fig/vit_small}};

\draw[{Circle[warmlight!80,length=4pt]}-{Circle[warmlight!80,length=4pt]},warmlight!80,very thick,rounded corners=3pt] ($(PE.south west)!.5!(PE.south east)$) |- ([xshift=-6pt]$(PE.north)!1.2!(PE.south)$) -| ([yshift=40pt]$(V.west)!.145!(V.east)$);
\draw[{Circle[cooldark!80,length=4pt]}-{Circle[cooldark!80,length=4pt]},cooldark!80,very thick,rounded corners=3pt] ($(FFN.south west)!.5!(FFN.south east)$) |- ([xshift=-6pt]$(FFN.north)!1.25!(FFN.south)$) -| ([yshift=28pt]$(V.west)!.7!(V.east)$);
\node[inner sep=0,anchor=south west] (LV) at ([yshift=2pt]V.north west) {B.~pixel-wise vision transformer};
\draw[{Circle[warmdark!80,length=4pt]}-{Circle[warmdark!80,length=4pt]},warmdark!80,very thick,rounded corners=3pt] (V.west) -| ([yshift=-6pt]$(V.east)!1.03!(V.west)$) |- ([xshift=24pt]$(U.south)!.06!(U.north)$);

\node[
    anchor=north west,
    font=\fontsize{7}{7}\selectfont, 
    inner sep=2pt, rounded corners, fill=white] (L) at ($(FFN.north west)!1.05!(FFN.north east)$) {
        \begin{tikzpicture}
            \def\bx{3pt}
            \def\by{5pt}
            \tikzstyle{sb} = [block,minimum width=6*\bx,minimum height=1.5*\by, inner sep=0]
            \tikzstyle{basicblock} = [sb, draw=coollight!80, fill=coollight!60]
            \tikzstyle{downsampling} = [sb, draw=cooldark!80, preaction={fill, cooldark!40}, pattern=north east lines, pattern color=cooldark!80]
            \tikzstyle{upsampling} = [sb, draw=warmlight!80, preaction={fill, warmlight!40}, pattern=north west lines, pattern color=warmlight!80]
            \tikzstyle{processing} = [sb, draw=other, fill=other!80]
        
            \node[basicblock, anchor=north west] (LBshape) at (0, 0) {};
            \node[inner sep=0, anchor=west] (LBtext) at ([xshift=\bx]LBshape.east) {basic block};
            
            \node[downsampling, anchor=north] (LDshape) at ([yshift=-\by]LBshape.south) {};
            \node[inner sep=0, anchor=north west, align=left] (LDtext) at ([xshift=\bx]LDshape.north east) {down-sampling:\\\texttt{Conv}$(k=2,s=2)$};
            
            \node[upsampling,anchor=north east] (LUshape) at ([yshift=-\by]LDshape.east |- LDtext.south) {};
            \node[inner sep=0, anchor=north west, align=left] (LUtext) at ([xshift=\bx]LUshape.north east) {
                up-sampling:\\
                up-sample with\\
                scale $2$ and\\
                \texttt{Conv}$(k=3, p=1)$
            };
            
            \node[circle,inner sep=0pt, anchor=north east,fill=black!30,draw=black!30] (LCshape) at ([yshift=-\by]LUshape.east |- LUtext.south) {\footnotesize $+$};
            \node[inner sep=0, anchor=west, align=left] (LCtext) at ([xshift=\bx]LCshape.east) {addition};
            
            \node[circle,inner sep=0pt,anchor=north east,fill=black!30,draw=black!30] (LAshape) at ([yshift=-\by]LCshape.south east) {\footnotesize $\oplus$};
            \node[inner sep=0,anchor=north west, align=left] (LAtext) at ([xshift=\bx]LAshape.north east) {concatenation along\\feature dimension};
            
            \node[inner sep=2pt, anchor=north east,fill=black!30,draw=black!30, rounded corners] (PREshape) at ([yshift=-\by]LAshape.east |- LAtext.south) {pre-process}; 
            \node[inner sep=0,anchor=north west, align=left] (PREtext) at ([xshift=\bx]PREshape.north east) {\texttt{Conv}$(k=3, p=1)$\\\texttt{LeakyReLU}};
            
            \node[inner sep=2pt, anchor=north east,fill=black!30,draw=black!30, rounded corners] (POSTshape) at ([yshift=-\by]PREshape.east |- PREtext.south) {post-process}; 
            \node[inner sep=0,anchor=north west, align=left] (POSTtext) at ([xshift=\bx]POSTshape.north east) {$1\times1$-\texttt{Conv}\\\texttt{sigmoid}};
            
            \node[fill=cooldark!40,rounded corners=1pt,anchor=north east] (ReZeroShape) at ([yshift=-\by]POSTshape.east |- POSTtext.south) {$\times\alpha$};
            \node[inner sep=0,anchor=north west, align=left] (ReZerotest) at ([xshift=\bx]ReZeroShape.north east) {learnable\\rezero parameter};
        \end{tikzpicture}
    };
\end{tikzpicture}

};
        \end{tikzpicture}
    }
    \caption{
    \textbf{Schematic diagrams of \vitgan.} A.~\uv generator; B.~pixel-wise ViT; C.~basic block; D.~positional embedding (PE); F.~feed-forward network (FFN).}
    \label{fig:gen}
\end{figure*}


A \uv generator consists of a UNet~\cite{ronneberger2015u} with
a pixel-wise Vision Transformer (ViT) \cite{dosovitskiy2020image} at the bottleneck (\autoref{fig:gen}A). 
UNet's encoding path extracts features from the input via four layers of convolution and downsampling. 
The features extracted at each layer are also passed to the corresponding layers of the decoding path via skip connections, whereas the bottom-most features are passed to the ViT.
We hypothesize that the skip connections are effective in passing high-frequency features to the decoder, and the ViT provides an effective means to learn pairwise relationships of low-frequency features.


On the encoding path of the UNet, the pre-processing layer turns an image into a tensor with dimension $(w_0,h_0,f_0)$. 
A pre-processed tensor will have its width and height halved at each down-sampling block, while the feature dimension doubled at the last three down-sampling blocks.  
The output from the encoding path with dimension $(w,h,f) = (w_0/16, h_0/16, 8f_0)$ forms the input to the pixel-wise ViT bottleneck. 

A pixel-wise ViT (\autoref{fig:gen}B) is composed primarily of a stack of Transformer encoder blocks~\cite{devlin2019bert}.
To construct an input to the stack, the ViT first flattens an encoded image along the spatial dimensions to form a sequence of tokens. 
The token sequence has length $w \times h$, and each token in the sequence is a vector of length $f$. 
It then concatenates each token with its two-dimensional Fourier positional embedding~\cite{anokhin2021image} of dimension $f_p$ (\autoref{fig:gen}D) and linearly maps the result to have dimension $f_v$. 
To improve the Transformer convergence, we adopt the rezero regularization~\cite{bachlechner2021rezero} scheme and introduce a trainable scaling parameter $\alpha$ that modulates the magnitudes of the nontrivial branches of the residual blocks. 
The output from the Transformer stack is linearly projected back to have dimension $f$ and unflattened to have width $w$ and $h$.
In this study, we use $12$ Transform encoder blocks and set $f, f_p, f_v=384$, and $f_h = 4f_v$ for the feed-forward network in each block (\autoref{fig:gen}E).


\subsection{Discriminator Loss with Gradient Penalty (GP)}
\label{subsec:gradientPenalty}
In this study, we use the least squares GAN (LSGAN) loss function~\cite{mao_least_2017} (i.e.,~$\ell_{\text{GAN}}$ is an $L_2$ error) in Eq.~\eqref{eq:disc_loss_A}-\eqref{eq:gen_loss_B} and supplement the discriminator loss with a GP term.
GP~\cite{gulrajani2017improved} originally was introduced to be used with Wasserstein GAN (WGAN) loss to ensure the $1$-Lipschitz constraint~\cite{arjovsky2017wasserstein}.
However, in our experiments, WGAN $+$ GP yielded overall worse results, which echoes the findings in~\cite{mao_effective_lsgan,mescheder2018training} 
We have also considered zero-centered GP~\cite{mescheder2018training}. In our case, zero-centered GP turned out to be very sensitive to the values of hyperparameters, and did not improve the training stability. Therefore, we settle on a more generic GP form introduced in~\cite{karras_progressive_2018} with the following formula for loss of $\mathcal{D}_A$:
\begin{equation}\label{eq:dist_loss_gp}
  \mathcal{L}^{\text{GP}}_{\text{disc}, A} = \mathcal{L}_{\text{disc}, A}
   + \lambda_\text{GP} \mathbb{E} \left[\frac{
      \left( \| \nabla_x \mathcal{D}_A(x) \|_2 - \gamma \right)^2}{\gamma^2}
    \right],
\end{equation}
where $\mathcal{L}_{\text{disc}, A}$ is defined as in Eq.~\eqref{eq:disc_loss_A}, and 
$\mathcal{L}^{\text{GP}}_{\text{disc}, B}$ follows the same form.
In our experiments, this $\gamma$-centered GP regularization provides more stable training and less sensitive to the hyperparameter choices.
To see the effect of GP on model performance, refer to the ablation study detailed in Section~\ref{subsec:pretraining_gradient-penalty} and Appendix Section~1.

\subsection{Self-Supervised Pre-training by Inpainting}
\label{subsec:pretraining}

Pre-training is an effective way to prime large networks for downstream tasks~\cite{devlin2019bert,bao_beit_2021} that often can bring significant improvement over random initialization.
In this work, we pre-train the \vitgan generators on an image inpainting task.
More precisely, we tile images with non-overlapping patches of size $32 \times 32$ and mask $40\%$ of the patches by setting their pixel values to zero.
The generator is trained to predict the original unmasked image using pixel-wise $L_1$ loss.
We consider two modes of pre-training: 1) on the same dataset where the subsequent image translation is to be performed and 2) on the ImageNet~\cite{5206848} dataset.
In Section~\ref{subsec:pretraining_gradient-penalty}, we conduct an ablation study on these two pre-training modes together with no pre-training.

\section{Experiments}
\subsection{Benchmarking Datasets}
\label{sec:datasets}


To test \vitgan's performance, we have completed an extensive literature survey for benchmark datasets. 
The most popular among them are datasets derived from CelebA~\cite{liu2015faceattributes} and Flickr-Faces~\cite{FFHQ}, as well as the SYNTHIA/GTA-to-Cityscape~\cite{ros2016synthia,cordts2016cityscapes, richter2016playing}, photo-to-painting~\cite{zhu_unpaired_2017}, \anime~\cite{kim_u-gat-it_2020}, and animal face datasets~\cite{choi2020starganv2}.
We prioritize our effort on the \anime dataset and two others derived from the CelebA dataset: gender swap (denoted as \gender) and adding and removing eyeglasses (marked as \glasses), which have been used in recent papers. 


\anime~\cite{kim_u-gat-it_2020} is a 
small dataset with 3.4K images in each domain.
Both \gender and \glasses tasks are derived from CelebA~\cite{liu2015faceattributes} based on the gender and eyeglass attributes, respectively.
\gender contains about 68K males and 95K females for training, while \glasses includes 11K with glasses and 152K without. 
For a fair comparison, we do not use CelebA's validation dataset for training. 
Instead, we combine it with the test dataset following the convention of~\cite{nizan_breaking_2019,zhao2020unpaired}.
\anime contains images of size $256 \times 256$ that can be used directly.
The CelebA datasets contains images of size $178\times 218$, which we resize and crop to size $256 \times 256$ for \vitgan training.
\subsection{UVCGAN Training Procedures}
\noindent\textbf{Pre-training.} The \vitgan generators are pre-trained with self-supervised image inpainting.  
To construct impaired images, we tile images of size $256 \times 256$ into non-overlapping $32 \times 32$ pixel patches and randomly mask $40\%$ of the patches by zeroing their pixel values.
We use the Adam optimizer, cosine annealing learning-rate scheduler, and several standard data augmentations, such as small-angle random rotation, random cropping, random flipping, and color jittering.
During pre-training, we do not distinguish the image domains, which means both generators in the ensuing translation training have the same initialization.
In this work, we pre-train one generator on ImageNet, another on CelebA, and one on the \anime dataset. 

\noindent\textbf{Image Translation Training.} For all three benchmarking tasks, we train \vitgan models for one million iterations with a batch size of one. 
We use the Adam optimizer with the learning rate kept constant at $0.0001$ during the first half of the training then linearly annealed to zero during the second half.
We apply three data augmentations: resizing, random cropping, and random horizontal flipping. 
Before randomly cropping images to $256 \times 256$, we enlarge them from $256 \times 256$ to $286 \times 286$ for \anime and $178 \times 218$ to $256 \times 313$ for CelebA. 

\noindent\textbf{Hyperparameter search.} 
The \vitgan loss functions depend on four hyperparameters:  $\lambda_{\text{cyc}}$, $\lambda_{\text{GP}}$, $\lambda_\text{idt}$ and $\gamma$, Eq.~\eqref{eq:gen_loss_A}-\eqref{eq:dist_loss_gp}.
If identity loss ($\lambda_\text{idt}$) is used, it is always set to $\lambda_\text{cyc} / 2$ as suggested in~\cite{zhu_unpaired_2017}. 
To find the best-performing configuration, we run a small-scale hyperparameter optimization on a grid.
Our experiments show that the best performance for all three benchmarking tasks is achieved with the LSGAN $+$ GP with $(\lambda_{\text{GP}} = 0.1, \gamma = 100)$ and with generators pre-trained on the image translation dataset itself. 
Optimal $\lambda_\text{cyc}$ differs slightly for CelebA and \anime at $5$ and $10$, respectively. 
An ablation study on hyperparameter tuning can be found in Section~\ref{subsec:pretraining_gradient-penalty}.
More training details also can be found in the open-source repository~\cite{github-uvcgan}.
\subsection{Other Model Training Details}
\label{subsec:train_benchmarking}


To fairly represent other models' performance, we strive to reproduce trained models following three principles.
First, if a pre-trained model for a dataset exists, we will use it directly. 
Second, in the absence of pre-trained models, we will train the model from scratch using a configuration file (if provided), following a description in the original paper, or using a hyperparameter configuration for a similar task.
Third, we will keep the source code ``as is'' unless it absolutely is necessary to make changes.
In addition, we have conducted a small-scale hyperparameter tuning on models lacking hyperparameters for certain translation directions (Appendix Sec.~2).
Post-processing and evaluation choices also will affect the reported performance (Section~\ref{subsec:reproducibility}).

\textbf{\acl}~\cite{github-aclgan} provides configuration file for the \gender dataset. 
For configuration files for \glasses and \anime, we copy the settings for \gender except for the four key parameters $\lambda_\text{acl}$, $\lambda_\text{mask}$, $\delta_\text{min}$, and $\delta_\text{max}$, which we modify according to the paper \cite[p.~8, Training Details]{zhao2020unpaired}.
Because \acl does not train two generators jointly, we train a model for each direction for all datasets. 
\textbf{\cou}~\cite{github-councilgan} provides models for all datasets but only in one direction (selfie to anime, male to female, removing glasses). 
The pre-tained models output images with size $256$ for \gender and \anime and $128$ for \glasses.
For a complete comparison, we train models for the missing directions using the same hyperparameters as the existing ones with the exception for \glasses -- we train a model for adding glasses for image size $256$.
\textbf{\cyc}~\cite{github-cyclegan} models are trained from scratch with the default settings (\texttt{resnet\_9blocks} generators and LSGAN losses, batch size $1$, etc.). 
Because the original \cyc uses square images, we add a pre-processing for CelebA by scaling up the shorter edge to $256$ while maintaining the aspect ratio, followed by a $256 \times 256$ random cropping. 
\textbf{\uga}~\cite{github-ugatit} provides the pre-trained model for \anime, which is used directly.
For the two CelebA datasets, models are trained using default hyperparameters.

\begin{table}[ht]
    \caption{\textbf{Training time}. \cyc, \uga, and \vitgan train two generators jointly. \acl and \cou's generators are trained separately for each direction. The time shown is for training both directions.}
    \label{tab:training_time}
    \setlength\tabcolsep{4pt}
    \addtolength{\extrarowheight}{\belowrulesep}
    \aboverulesep=0pt
    \belowrulesep=0pt
    \newcolumntype{g}{>{\columncolor{black!10}}r}
    \centering
    \begin{tabular}{r|g|c|l}
        \toprule
        \rowcolor{black!30}
        Algorithm & Time (hrs) & Jointly Trained & \# Para.\\
        \midrule
        \acl &      $\sim86$ & & $\sim55$M\\
        \cou &     $\sim600$ & & $\sim116$M\\  
        \cyc &      $\sim40$ & \checkmark & $\sim28$M\\
        \uga &     $\sim140$ & \checkmark & $\sim671$M\\
        \vitgan &   $\sim60$ & \checkmark & $\sim68$M\\
        \bottomrule
    \end{tabular}
\end{table}

\autoref{tab:training_time} depicts the training time (in hours) for various models on the CelebA datasets using an NVIDIA RTX A6000 GPU.
The times correspond to training the models with a batch size $1$ for one million iterations.
\uga's  long training time is due to a large number of loss function terms that must be computed, as well as the large size of the generators and discriminators. For \cou, the time stems from training an ensemble of generators, each with its own discriminator, in addition to the domain discriminators.
More details are available in the open-source repository~\cite{github-uvcgan-benchmarking}


\section{Results}
\def\numImages{8}
\begin{figure*}[ht]
\centering
\includegraphics[width=\textwidth]{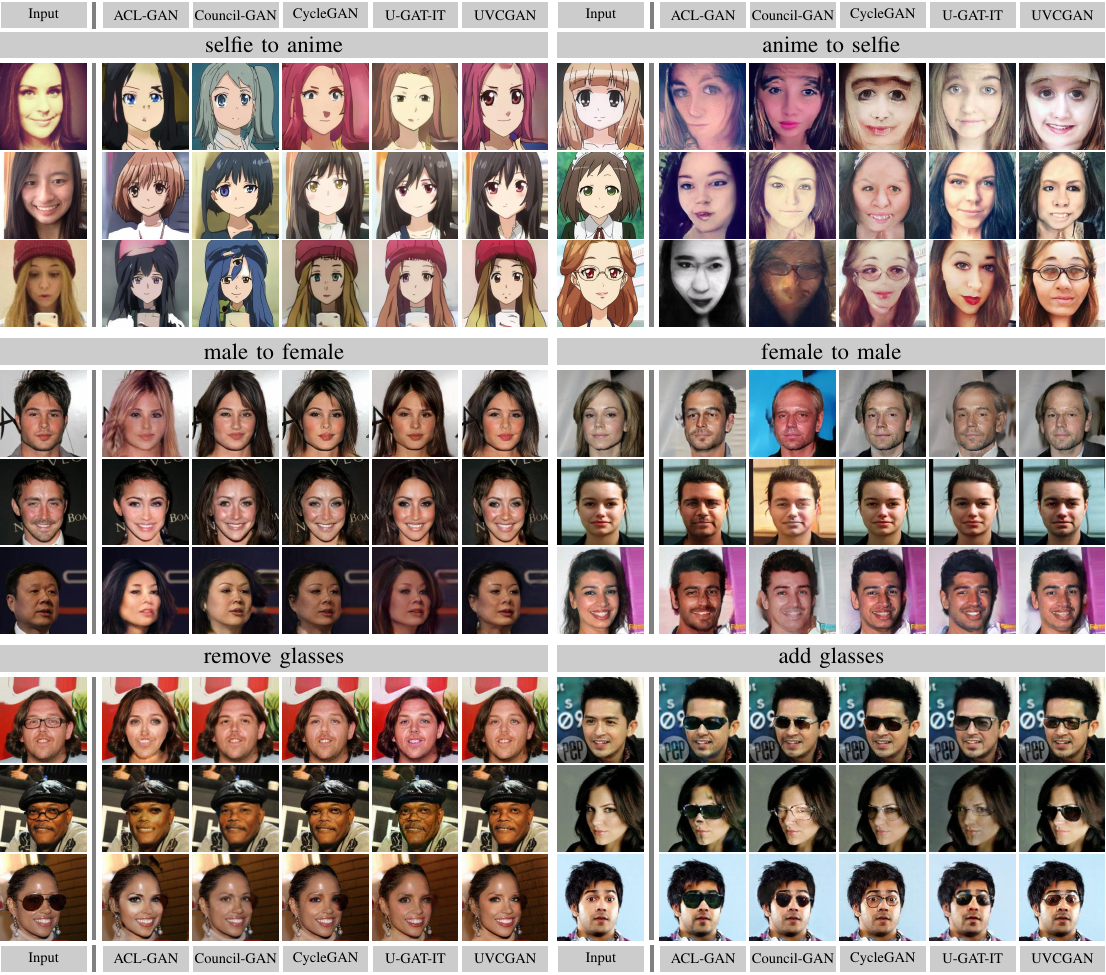}
\caption{\textbf{Samples of unpaired image-to-image translation. } 
}
\label{fig:image_grid}



\label{fig:img_glasses}
\end{figure*}
\subsection{Evaluation Metrics}
\label{sec:scores}

Fr\'{e}chet Inception Distance (FID)~\cite{NIPS2017_8a1d6947} and Kernel Inception Distance (KID)~\cite{binkowski2018demystifying}
are the two most accepted metrics used for evaluating image-to-image translation performance.
A lower score means the translated images are more similar to those in the target domain.
As shown in \autoref{tab:results}, our model offers better performance in most image-to-image translation tasks
compared to existing models.
As a \cyc-like model, ours produce translated images correlated strongly with the input images, 
such as hair color and face orientations (\autoref{fig:image_grid}),
which is crucial for augmenting scientific simulations.
On the contrary, we observed the translations produced by \acl and \cou tend to be overly liberal on features that are not essential in accomplishing the translation (such as background color or hair color and length). 
We also note that although \uga achieves lower scores in the anime-to-selfie task and produces translations that resemble human faces better, they are less correlated to the input and sometimes miss key features from the input (such as headdress or glasses) that we want to preserve. 
In the Supplementary material, more samples of larger sizes are provided.


\begin{table}[!ht]
    \caption{\textbf{FID and KID scores.} Lower is better.
    }
    \label{tab:results}
    \setlength\tabcolsep{4pt}
    \addtolength{\extrarowheight}{\belowrulesep}
    \aboverulesep=0pt
    \belowrulesep=0pt
    \centering
    
    \newcolumntype{g}{>{\columncolor{black!10}}r}
    \newcolumntype{i}{>{\columncolor{black!30}}r}
    \def\lcolrule{\color{black!0}\vrule}
    \def\pcolrule{\color{black!0}\vrule}
    
    \resizebox{\linewidth}{!}{
    \begin{threeparttable}
        \begin{tabular}{i|r!{\pcolrule}g!{\lcolrule}r!{\pcolrule}g}
            \toprule
                \rowcolor{black!30}
                 & \multicolumn{2}{r!{\color{black!20}\vrule}}{Selfie to Anime} & \multicolumn{2}{r}{Anime to Selfie} \\
                 & FID & KID ($\times100$) & FID & KID ($\times100$) \\ 
            \acl & $99.3$ & $3.22\pm0.26$ & $128.6$ & $3.49\pm0.33$ \\
            \cou & \underline{$91.9$} & $2.74\pm0.26$ & $126.0$ & $2.57\pm0.32$ \\
            \cyc & $92.1$ & \underline{$2.72\pm0.29$} & $127.5$ & $2.52\pm0.34$ \\
            \uga & $95.8$ & $2.74\pm0.31$ & $\mathbf{108.8}$ &  $\mathbf{1.48\pm0.34}$ \\
            \arrayrulecolor{black!75}\midrule
            \vitgan & $\mathbf{79.0}$ & $\mathbf{1.35\pm0.20}$ & \underline{$122.8$} &  \underline{$2.33\pm0.38$} \\
            \arrayrulecolor{black!75}\midrule
                \rowcolor{black!30}
                 & \multicolumn{2}{r!{\color{black!20}\vrule}}{Male to Female} & \multicolumn{2}{r}{Female to Male} \\
                 & FID & KID ($\times100$) & FID & KID ($\times100$) \\ 
            \acl & $\mathbf{9.4}$ & $\mathbf{0.58\pm0.06}$ & $19.1$ & $1.38\pm0.09$ \\
            \cou & $10.4$ & $0.74\pm0.08$ & $24.1$ & $1.79\pm0.10$ \\
            \cyc  & $15.2$ & $1.29\pm0.11$ & $22.2$ & $1.74\pm0.11$ \\
            \uga & $24.1$ & $2.20\pm0.12$ & \underline{$15.5$} & \underline{$0.94\pm0.07$} \\
            \arrayrulecolor{black!75}\midrule 
            \vitgan & \underline{$9.6$} & \underline{$0.68\pm0.07$} & $\mathbf{13.9}$ & $\mathbf{0.91\pm0.08}$ \\ 
            \arrayrulecolor{black!75}\midrule
                \rowcolor{black!30}
                 & \multicolumn{2}{r!{\color{black!20}\vrule}}{Remove Glasses} & \multicolumn{2}{r}{Add Glasses} \\
                 & FID & KID ($\times100$) & FID & KID ($\times100$) \\ 
            \acl & \underline{$16.7$} & \underline{$0.70\pm0.06$} & $20.1$ & $1.35\pm0.14$ \\
            \cou & $37.2$ & $3.67\pm0.22$ &    $19.5$ & $1.33\pm0.13$ \\
            \cyc & $24.2$ & $1.87\pm0.17$ & $19.8$ & $1.36\pm0.12$ \\
            \uga & $23.3$ & $1.69\pm0.14$ & \underline{$19.0$} & \underline{$1.08\pm0.10$} \\
            \arrayrulecolor{black!75}\midrule  
            \vitgan & $\mathbf{14.4}$ & $\mathbf{0.68\pm0.10}$ & $\mathbf{13.6}$ & $\mathbf{0.60\pm0.08}$ \\
            \arrayrulecolor{black}\bottomrule
        \end{tabular}
    \end{threeparttable}
    }
\end{table}


\subsection{Model Evaluation and Reproducibility}
\label{subsec:reproducibility}
KID and FID for image-to-image translation are difficult to reproduce.
For example, in~\cite{nizan_breaking_2019,zhao2020unpaired,kim_u-gat-it_2020}, 
most FID and KID scores of the same task-model settings differ.
We hypothesize that this is due to: 
1) Different sizes of test data as FID decreases with more data samples~\cite{binkowski2018demystifying} 
2) Differences in post-processing before testing 
3) Different formulation of metrics (e.g. KID in \uga~\cite{kim_u-gat-it_2020})
4) Different FID and KID implementations.
Therefore, we standardize the evaluations as follows: 1) Using the full test datasets for FID and KID---for KID subset size, use $50$ for \anime and $1000$ for the two CelebA datasets; 2) Resizing non-square CelebA images and taking a central crop of size $256 \times 256$ to maintain the correct aspect ratio; 3) Delegating all KID and FID calculations to the torch-fidelity package~\cite{obukhov2020torchfidelity}.

\textbf{\acl} follows a non-deterministic type of cycle consistency and can generate a variable number of translated images for an input. However, because larger sample size improves FID score~\cite{binkowski2018demystifying}, we generate one translated image per input for a fair comparison.
To produce the test result, \acl resizes images from CelebA to have width $256$ and output without cropping.
For FID and KID evaluation, we take the center $256 \times 256$ crop from the test output.
\textbf{\cou} resizes the images to have a width $256$, except for removing glasses, which is $128$ due to the pre-trained model provided. 
In order to follow the principle of using a pre-trained model if available and maintain consistency in evaluating on images of size $256$, we resize $128$ to $256$ during testing, which may be responsible for the large FID score.
The reverse direction, adding glasses, is trained from scratch using an image size of $256$. Its performance is similar to that of other models.
\textbf{\cyc} takes a random square crop for both training and testing.
However, for a fair comparison, we modify the source code so the test output are the center crops.
Since the original \textbf{\uga} cannot handle non-square images, we modified the code to scale the shorter edge $256$ for the CelebA datasets. 

\subsection{Ablation Studies}
\label{subsec:pretraining_gradient-penalty}

\autoref{table:ablation_study} summarizes the male-to-female and
selfie-to-anime translation performance with respect to 
pre-training, 
GP, and identity loss. 
First, GP combined with identity loss improves performance. 
Second, without GP, the identity loss produces mixed results.
Finally, pre-training on the same dataset improves performance, especially in conjunction with the GP and identity loss. 
Appendix Sec.~1 contains the complete ablation study results for all data sets.

We speculate that the GP is required to obtain the best performance with pre-trained networks because those networks provide a good starting point for the image translation task. However, at the beginning of fine-tuning, the discriminator is
initialized by random values and provides a meaningless signal to the
generator. This random signal may drive the generator away from the good
starting point and undermine the benefits of pre-training. 


\begin{table}[ht]
\caption{\label{table:ablation_study}\textbf{Ablation studies.} 
Pre-train/Dataset column indicates which dataset the generator is pre-trained on (\textit{None} for no pre-training; \textit{Same} indicates CelebA for male-to-female and Selfie2Anime for selfie-to-anime).}
  \setlength\tabcolsep{4pt}
  \addtolength{\extrarowheight}{\belowrulesep}
  \aboverulesep=0pt
  \belowrulesep=0pt
  \centering
  \small
  
  \newcolumntype{i}{>{\columncolor{black!30}}r}
  \newcolumntype{g}{>{\columncolor{black!10}}r}
\resizebox{\linewidth}{!}{
\begin{tabular}{i|c|c|rg|rg}
\toprule
\rowcolor{black!30}Pre-train & & & \multicolumn{2}{c|}{Male to Female} & \multicolumn{2}{c}{Selfie to Anime} \\
Dataset  & GP & $\text{idt}$ & FID & KID ($\times 100$) & FID & KID ($\times 100$) \\
\midrule
Same     & \checkmark & \checkmark & $9.6$  & $0.68 \pm 0.07$ & $79.0$ & $1.35 \pm 0.20$ \\
ImageNet & \checkmark & \checkmark & $11.0$ & $0.85 \pm 0.08$ & $81.3$ & $1.66 \pm 0.21$ \\
None     & \checkmark & \checkmark & $11.0$ & $0.85 \pm 0.09$ & $80.9$ & $1.78 \pm 0.20$ \\
\midrule
Same     & \checkmark &            & $11.1$ & $0.86 \pm 0.08$ & $83.9$ & $1.88 \pm 0.35$ \\
ImageNet & \checkmark &            & $11.0$ & $0.85 \pm 0.08$ & $84.3$ & $1.77 \pm 0.21$ \\
None     & \checkmark &            & $13.4$ & $1.11 \pm 0.09$ & $115.4$ & $6.85 \pm 0.59$ \\
\midrule
Same     &            & \checkmark & $14.2$ & $1.22 \pm 0.10$ & $81.5$ & $1.68 \pm 0.22$ \\
ImageNet &            & \checkmark & $14.5$ & $1.23 \pm 0.10$ & $86.8$ & $2.21 \pm 0.25$ \\
None     &            & \checkmark & $14.4$ & $1.26 \pm 0.10$ & $81.6$ & $1.75 \pm 0.25$ \\
\midrule
Same     &            &            & $12.7$ & $1.06 \pm 0.09$ & $79.0$ & $1.32 \pm 0.19$ \\
ImageNet &            &            & $13.4$ & $1.14 \pm 0.10$ & $91.2$ & $2.63 \pm 0.23$ \\
None     &            &            & $18.3$ & $1.63 \pm 0.11$ & $81.2$ & $1.76 \pm 0.21$ \\
\bottomrule
\end{tabular}
}
\end{table}

\begin{figure*}[ht]
    \centering
    \tikzsetnextfilename{attention}
    \input{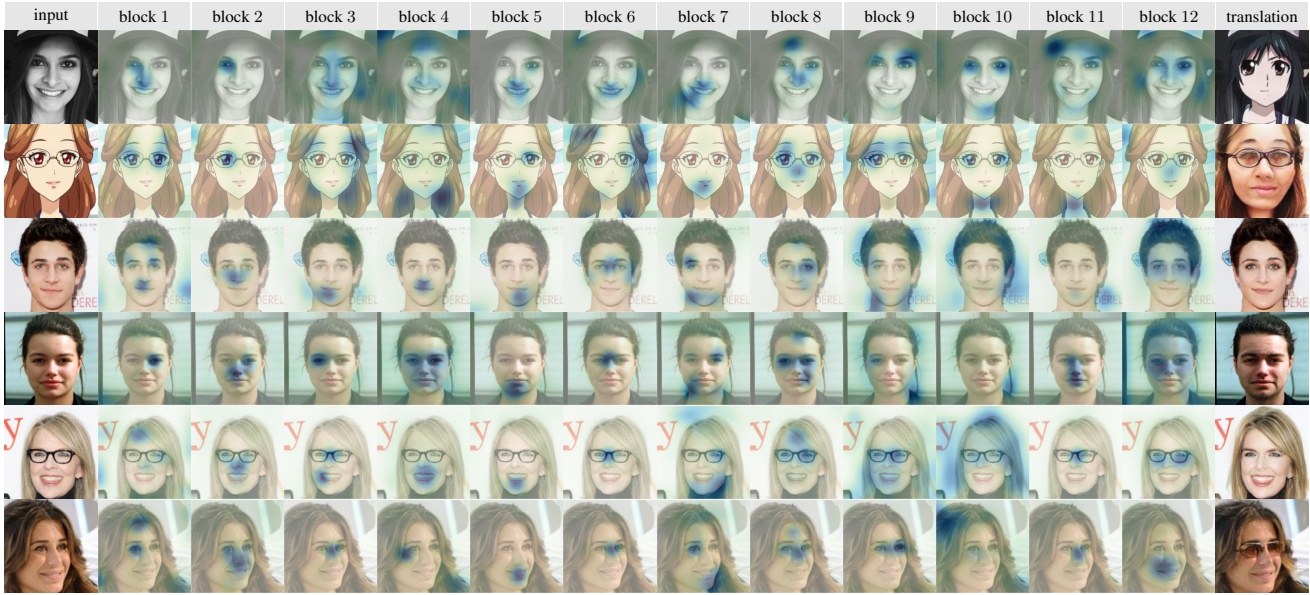}
    \caption{\textbf{Attention.} Attention heatmap generated by the attention weights from the 12 Transformer encoder blocks in the pixel-wise ViT. The attention heatmap demonstrates the amount of attention different locations of an image receive.}
    \label{fig:attention}
\end{figure*}

\subsection{Interpretation of Attention}
\label{subsec:attention}

Because the UVCGAN generator uses the transformer bottleneck, it is instructive to visualize its attention matrices to see if they help with generator interpretability. 
We plot (\autoref{fig:attention}) the attention weights produced by the multi-head self-attention (MSA) unit in each of the $12$ Transformer encoder blocks in the 
bottleneck of the \vitgan generators (\autoref{fig:gen}B). 
The $(i, j)$-entry of the attention matrix indicates how much attention token $i$ is paying to token $j$ while the sum of row $i$ is one. 
When multi-head attention is used, each head produces an attention matrix. 
For simplicity, we average the attention weights over all heads and target tokens for each block in the Transformer encoder stack.
Given the input image of size $256 \times 256$, this provides an attention vector of dimension $w \times h$ ($16 \times 16=256$). 
The $j$-th entry of such a vector indicates how much attention token $j$ receives on average. 
Because the tokens represent overlapping patches in the original image, we generate a heatmap as follows: reshape a feature vector to a square of size $16 \times 16$, upscale it 16 folds to match the dimension of the input image, then apply a Gaussian filter with $\sigma=16$. 
By overlaying the attention heatmap on the input images, we note that each block is paying attention to a specific facial part with the eye and mouth areas receiving the most attention. 
This echoes the findings in behavioral science experiments on statistical eye fixation (e.g., ~\cite{caldara2011imap}), where the regions of interest also tend to be around the eyes and mouth,
which may indicate that the model's attention is focused on the most informative and relevant regions.

\section{Conclusion}

This work introduces \vitgan to promote cycle-consistent, content-preserving image
translation and effectively handle long-range spatial dependencies that remain
a common problem in scientific domain research. 
Combined with self-supervised pre-training and GP regularization, \vitgan outperforms competing methods on a diverse set of image translation benchmarks. 
The ablation study suggests GP and cycle-consistent loss work well with \vitgan.
Additional inspection on attention weights indicates our model
has focused on the relevant regions of the source images.
To further demonstrate the effectiveness of our model in handling long-distance patterns beyond benchmark datasets, more open scientific datasets are needed. 

\noindent\textbf{Potential Negative Social Impact.}  
All data used in this work are publicly available. 
The environmental impact of training our model is greater than the original \cyc yet considerably less comparing to other advanced models. 
Although the motivation of our image-to-image translation work is to bridge the gap between scientific simulation and experiment,
the authors are aware of its potential use for generating fake content~\cite{mirsky2021creation}.
Thankfully, there are countermeasures and detection tools~\cite{juefei2022countering} developed to counter such misuse.
To contribute to such mitigation efforts, we have made our code and pre-trained models available. 

\noindent\textbf{Acknowledgement.}
The LDRD Program at Brookhaven National Laboratory, sponsored by DOE’s Office of Science under Contract DE-SC0012704, supported this work.

\newpage
{
\small
\bibliographystyle{ieee_fullname}
\bibliography{bib/bib}
}

\newpage
\begin{appendices}
\setcounter{table}{0}
\setcounter{figure}{0}
\renewcommand{\thetable}{A\arabic{table}}
\renewcommand{\thefigure}{A\arabic{figure}}
\renewcommand{\figurename}{Figure}
\renewcommand{\tablename}{Table}
\section{Extended \vitgan Ablation Studies}

This appendix shows the impact of the \vitgan generator, gradient penalty (\textbf{GP}), and self-supervised generator pretraining (\textbf{PT}) on \vitgan's performance. \autoref{tab:results_extended} summarizes these findings.
For each data set, the bottom half of the table shows the \vitgan performance with some of its components disabled.
For example, \textit{\vitgan no GP} shows the \vitgan performance without the gradient penalty term (but using a hybrid \uv generator and a self-supervised pretraining). This table affords a few observations:
1.~the addition of a hybrid \uv generator alone typically produces a large degree of improvement compared to \cyc, even in the absence of the self-supervised pre-training and GP term;
2.~the self-supervised generator pre-training without the GP term does not seem to improve the image-to-image translation performance and sometimes makes it worse;
3.~the self-supervised pre-training only helps when it is used in conjunction with the GP.

\begin{table}[!ht]
    \caption{\textbf{FID and KID scores.} Lower is better. \textbf{PT} stands for the self-supervised generator pre-training, and \textbf{GP} means usage of the gradient penalty.
    }
    \label{tab:results_extended}
    \setlength\tabcolsep{4pt}
    \addtolength{\extrarowheight}{\belowrulesep}
    \aboverulesep=0pt
    \belowrulesep=0pt
    \centering
    
    \newcolumntype{g}{>{\columncolor{black!10}}r}
    \newcolumntype{i}{>{\columncolor{black!30}}r}
    \def\lcolrule{\color{black!0}\vrule}
    \def\pcolrule{\color{black!0}\vrule}
    
    \resizebox{\linewidth}{!}{
    \begin{threeparttable}
        \begin{tabular}{i|r!{\pcolrule}g!{\lcolrule}r!{\pcolrule}g}
            \toprule
                \rowcolor{black!30}
                 & \multicolumn{2}{r!{\color{black!20}\vrule}}{Selfie to Anime} & \multicolumn{2}{r}{Anime to Selfie} \\
                 & FID & KID ($\times100$) & FID & KID ($\times100$) \\ 
            \acl & $99.3$ & $3.22\pm0.26$ & $128.6$ & $3.49\pm0.33$ \\
            \cou & $91.9$ & $2.74\pm0.26$ & $126.0$ & $2.57\pm0.32$ \\
            \cyc & $92.1$ & $2.74\pm0.31$ & $127.5$ & $2.52\pm0.34$ \\
            \uga & $95.8$ & $2.74\pm0.31$ & $\mathbf{108.8}$ &  $\mathbf{1.48\pm0.34}$ \\
            \arrayrulecolor{black!75}\midrule
            \vitgan & $\mathbf{79.0}$ & $\mathbf{1.35\pm0.20}$ & \underline{$122.8$} &  \underline{$2.33\pm0.38$} \\
            \vitgan no GP  & $81.4$ & \underline{$1.68\pm0.22$} & $133.3$ &  $2.90\pm0.49$ \\
            \vitgan no PT  & \underline{$80.9$} & $1.78\pm0.20$ & $134.0$ &  $2.98\pm0.49$ \\
            \vitgan no PT and GP & $81.6$ & $1.75\pm0.25$ & $140.6$ &  $3.53\pm0.59$ \\
            \arrayrulecolor{black!75}\midrule
                \rowcolor{black!30}
                 & \multicolumn{2}{r!{\color{black!20}\vrule}}{Male to Female} & \multicolumn{2}{r}{Female to Male} \\
                 & FID & KID ($\times100$) & FID & KID ($\times100$) \\ 
            \acl & $\mathbf{9.4}$ & $\mathbf{0.58\pm0.06}$ & $19.1$ & $1.38\pm0.09$ \\
            \cou & $10.4$ & $0.74\pm0.08$ & $24.1$ & $1.79\pm0.10$ \\
            \cyc  & $15.2$ & $1.29\pm0.11$ & $22.2$ & $1.74\pm0.11$ \\
            \uga & $24.1$ & $2.20\pm0.12$ & $15.5$ & $0.94\pm0.07$ \\
            \arrayrulecolor{black!75}\midrule 
            \vitgan & \underline{$9.6$} & \underline{$0.68\pm0.07$} & $\mathbf{13.9}$ & $\mathbf{0.91\pm0.08}$ \\ 
            \vitgan no GP  & $14.1$ & $1.22\pm0.10$ & $20.4$ &  $1.61\pm0.11$ \\
            \vitgan no PT  & $11.0$ & $0.85\pm0.09$ & \underline{$14.7$} &  \underline{$0.98\pm0.08$} \\
            \vitgan no PT and GP & $14.4$ & $1.26\pm0.10$ & $19.9$ &  $1.55\pm0.11$ \\
            \arrayrulecolor{black!75}\midrule
                \rowcolor{black!30}
                 & \multicolumn{2}{r!{\color{black!20}\vrule}}{Remove Glasses} & \multicolumn{2}{r}{Add Glasses} \\
                 & FID & KID ($\times100$) & FID & KID ($\times100$) \\ 
            \acl & $16.7$ & \underline{$0.70\pm0.06$} & $20.1$ & $1.35\pm0.14$ \\
            \cou & $37.2$ & $3.67\pm0.22$ & $19.5$ & $1.33\pm0.13$ \\
            \cyc & $24.2$ & $1.87\pm0.17$ & $19.8$ & $1.36\pm0.12$ \\
            \uga & $23.3$ & $1.69\pm0.14$ & $19.0$ & $1.08\pm0.10$ \\
            \arrayrulecolor{black!75}\midrule  
            \vitgan & $\mathbf{14.4}$ & $\mathbf{0.68\pm0.10}$ & $\mathbf{13.6}$ & $\mathbf{0.60\pm0.08}$ \\
            \vitgan no GP  & $19.2$ & $1.28\pm0.15$ & $18.7$ &  $1.14\pm0.12$ \\
            \vitgan no PT  & \underline{$15.8$} & $0.84\pm0.12$ & \underline{$14.3$} &  \underline{$0.70\pm0.10$} \\
            \vitgan no PT and GP & $19.7$ & $1.32\pm0.15$ & $16.1$ &  $0.89\pm0.11$ \\
            \arrayrulecolor{black}\bottomrule
        \end{tabular}
    \end{threeparttable}
    }
\end{table}

\section{Hyperparameter Tuning for Other Algorithms}
This section summarizes the hyperparameter tuning results for three benchmarking algorithms: ACL-GAN, CycleGAN, and U-GAT-IT.
We omitted tuning for Council-GAN because it takes too long to run ($300$ hours per translation).

Because none of the benchmarking algorithms use any stablization techniques (such as the EMA of network weight~\cite{karras2020analyzing}) beyond shrinking learning rate, 
we suspect the fluctuation may be at least partially due to instability of the GAN training.

We only provide hyperparameter tuning results for a data set or task if an algorithm did not work on it.
We skip hyperparameter tuning if either a pre-trained model or a hyperparameter setup was provided by the author. 
In Table~\ref{tab:acl_hp}-\ref{tab:uga_hp}, the best results are marked in bold font.
The default hyperparameters are highlighted in \hl{gray}.

\textbf{ACL-GAN} worked on all three data sets studied and detailed in this paper---but all for only one direction: selfie-to-anime, male-to-female, and remove glasses.
For the translation in the opposite directions, we tune three parameters concerning the focus loss: focus loss weight, focus upper, and focus lower.
The results are summarized in Table~\ref{tab:acl_hp}.

\def\cg{\cellcolor{black!20}}
\begin{table}[ht]
    \centering
    \caption{\textbf{ACL-GAN hyperparameter tuning results.} We tune three hyperparameters related to the focus loss: weight of the focus loss, focus upper, and focus lower.}
    \resizebox{\linewidth}{!}{
    \begin{tabular}{l|r|r|r|c|c}
        \toprule
        task                             & weight     	& upper 	& lower  	& FID 				 	& KID($\times100$) \\ 
        \midrule
        \multirow{3}{*}{anime-to-selfie} & \cg $0$ 		& \cg $-$	& \cg $-$	& \cg $\mathbf{128.6}$  & \cg $\mathbf{3.49\pm0.33}$ \\
                                         & $.025$       & $.5$  	& $.3$   	& $205.3$ 		     	& $11.0\pm1.01$ \\
                                         & $.025$    	& $.1$  	& $.05$  	& $250.3$ 		     	& $18.6\pm1.19$ \\ 
        \midrule
        \multirow{3}{*}{female-to-male}  & $0$       	& $-$   	& $-$    	& $46.0$			 	& $3.39\pm0.13$ \\ 
                                         &\cg $.025$	& \cg $.5$  & \cg $.3$  & \cg $\mathbf{19.1}$  	& \cg $\mathbf{1.38\pm0.09}$ \\
                                         & $.05$     	& $.5$  	& $.3$   	& $36.3$ 		     	& $2.91\pm0.13$ \\ 
        \midrule
        \multirow{3}{*}{add glasses}     & $0$       	& $-$   	& $-$    	& $29.0$ 		     	& $1.77\pm0.12$ \\
                                         &\cg $.025$    & \cg $.1$  & \cg $.05$ & \cg $26.6$ 		    & \cg $2.26\pm0.17$ \\
                                         & $.05$    	& $.1$  	& $.05$  	& $\mathbf{20.1}$   	& $\mathbf{1.35\pm0.14}$ \\
        \bottomrule
    \end{tabular}
    }
    \label{tab:acl_hp}
\end{table}

\textbf{CycleGAN} did not work on any of the three data sets.
We search a grid on two hyperparameters: type of generator (Gen.) and weight (Wt.) of cycle-consistency loss. 
We also try two GAN modes: lsgan and wgangp. However, because CycleGAN did not implement GP properly, wgangp did not work.
The results are summarized in Table~\ref{tab:cyc_hp}.
\begin{table}[ht]
    \centering
    \caption{\textbf{CycGAN hyperparameter tuning results.}}
    \resizebox{\linewidth}{!}{
    \begin{tabular}{r|r|c|c|c|c}
        \toprule
                &        & FID & KID($\times100$) & FID & KID($\times100$) \\ \cline{3-6}
        gen.    & Wt.    & \multicolumn{2}{c|}{selfie-to-anime} & \multicolumn{2}{c}{anime-to-selfie} \\
        \midrule
        ResNet  & $5$    &  $\mathbf{92.1}$	& $\mathbf{2.72\pm0.29}$ & $\mathbf{127.5}$ &	$\mathbf{2.52\pm0.34}$ \\
        \rowcolor{black!20} ResNet  & $10$   &  $93.4$ & $2.96\pm0.27$ & $129.4$ & $2.91\pm0.39$ \\
        UNet    & $5$    &  $121.9$	& $6.21\pm0.32$ & $134.3$ &	$2.96\pm0.30$ \\
        UNet    & $10$   &  $286.0$	& $27.0\pm0.87$ & $135.8$ & $3.32\pm0.32$ \\ \midrule
                &        & \multicolumn{2}{c|}{male-to-female} & \multicolumn{2}{c}{female-to-male} \\ \midrule
        ResNet  & $5$    &  $21.9$ & $2.00\pm0.12$ & $33.6$ & $2.82\pm0.14$\\
        \rowcolor{black!20} ResNet  & $10$   &  $\mathbf{15.2}$ & $\mathbf{1.29\pm0.11}$ & $\mathbf{22.2}$ & $\mathbf{1.74\pm0.11}$ \\
        UNet    & $5$    &  $45.5$ & $4.55\pm0.17$ & $50.8$ & $4.86\pm0.16$\\
        UNet    & $10$   &  $47.4$ & $4.82\pm0.19$ & $47.5$ & $4.57\pm0.17$ \\ \midrule
                &        & \multicolumn{2}{c|}{remove glasses} & \multicolumn{2}{c}{add glasses} \\ \midrule
        ResNet  & $5$    & $27.7$ & $2.08\pm0.16$ & $26.0$ & $1.77\pm0.11$\\
        \rowcolor{black!20} ResNet  & $10$   &  $\mathbf{24.2}$ & $\mathbf{1.87\pm0.17}$ & $\mathbf{19.8}$ & $\mathbf{1.36\pm0.12}$ \\
        UNet    & $5$    & $32.2$ & $2.52\pm0.19$ & $37.3$ & $2.90\pm0.14$ \\
        UNet    & $10$   & $32.2$ & $2.52\pm0.19$ & $44.9$ & $3.63\pm0.20$ \\
        \bottomrule
    \end{tabular}
    }
    \label{tab:cyc_hp}
\end{table}

In addition to hyperparameter tuning for \textbf{U-GAT-IT}, we also correct the aspect ratio problem of U-GAT-IT in this revised version as the original U-GAT-IT implementation cannot handle images with different height and width.
We implement the rescaling in the preprocessing stage, so a CelebA image of width $178$ and height $218$
is resized to have width $256$ and height $313$. 
As we did for CycleGAN and UVCGAN, we take a random $256\times256$ crop from a training image and a central $256\times256$ crop from a test image. 

U-GAT-IT studied the selfie-to-anime data set. For the two CelebA data sets, we try three levels of weight of cycle-consistency loss: $(5, 10, \text{and }20)$ and summarize the results in Table~\ref{tab:uga_hp}.
\begin{table}[ht]
    \centering
    \caption{\textbf{U-GAT-IT hyperparameter tuning results.}}
    \resizebox{\linewidth}{!}{
    \begin{tabular}{r|c|c|c|c}
        \toprule
                & FID & KID($\times100$) & FID & KID($\times100$) \\ \cline{2-5}
        weight  & \multicolumn{2}{c|}{male-to-female} & \multicolumn{2}{c}{female-to-male} \\ \midrule
        $5$     &  $39.2$ &	$3.86\pm0.15$ & $45.1$ & $4.04\pm0.16$\\
        \rowcolor{black!20} $10$    &  $\mathbf{24.1}$ & $\mathbf{2.20\pm0.12}$ & $\mathbf{15.5}$ & $\mathbf{0.94\pm0.07}$\\
        $20$    &  $32.1$ & $3.09\pm0.16$ & $47.5$ & $4.42\pm0.17$\\ \midrule
                & \multicolumn{2}{c|}{remove glasses} & \multicolumn{2}{c}{add glasses} \\ \midrule
        $5$     &  $34.9$ & $2.63\pm0.15$ & $50.0$ & $5.08\pm0.26$\\
        \rowcolor{black!20} $10$    &  $\mathbf{23.3}$ & $\mathbf{1.69\pm0.14}$ & $\mathbf{19.0}$ & $\mathbf{1.08\pm0.10}$\\
        $20$    &  $36.1$ & $3.13\pm0.19$ & $36.1$ & $2.67\pm0.13$ \\
        \bottomrule
    \end{tabular}}
    \label{tab:uga_hp}
\end{table}
\section{More detail about the \uv Generator}

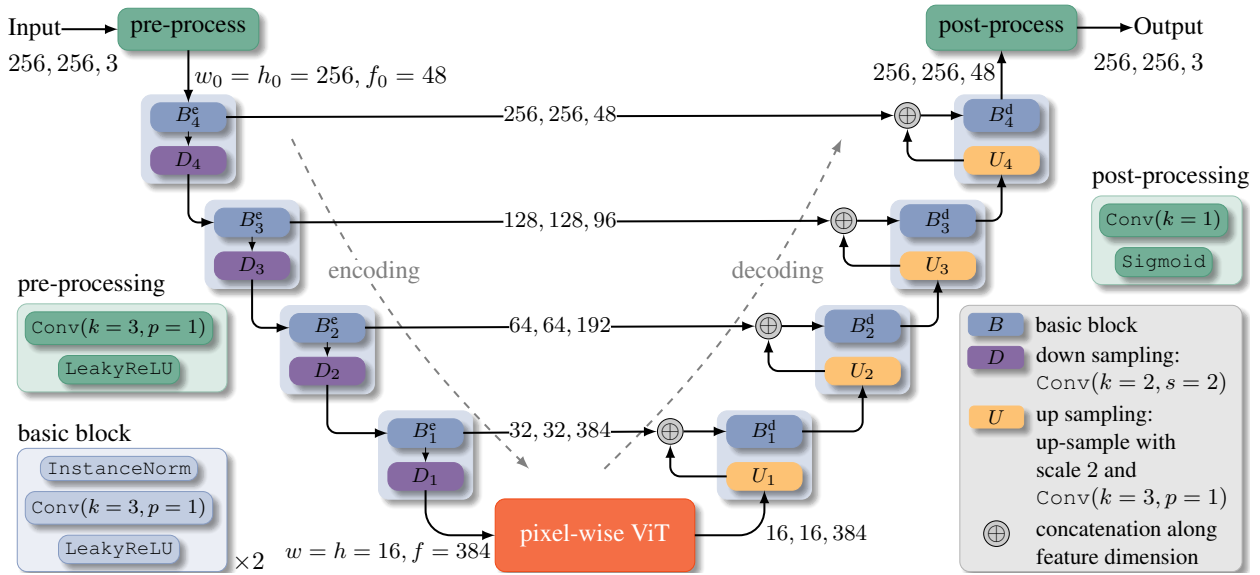
\begin{figure*}
    \centering
    \resizebox{.95\textwidth}{!}{
    \begin{tikzpicture}
        \node[inner sep=0, fill=white] at (0, 0) {


\begin{tikzpicture}
\tikzstyle{block} = [
    rectangle, 
    rounded corners, 
    blur shadow={shadow blur steps=5},
    align=center,
    draw=black, 
	fill=black!5,
    inner sep=5pt,
    outer sep=0pt,
]
\def\xsep{27pt}
\def\ysep{42pt}
\def\bx{15pt}
\def\by{5.5pt}

\def\levOne{3.5}
\def\levTwo{4.95}
\def\levThree{6.06}
\def\levFour{7}

\tikzstyle{encdec} = [block, draw=coollight!20, fill=coollight!20, inner xsep=1.25 * \bx, inner ysep=3.2 * \by]
\tikzstyle{bb} = [block,font=\fontsize{8}{8}\selectfont,inner xsep=.5*\bx,inner ysep=.3*\by,minimum height=2*\by,minimum width=2*\bx]
\tikzstyle{basicblock}   = [bb, draw=coollight!60, fill=coollight!60]
\tikzstyle{downsampling} = [bb, draw=cooldark!60,  fill=cooldark!60]
\tikzstyle{upsampling}   = [bb, draw=warmlight!60, fill=warmlight!60]
\tikzstyle{bottleneck}   = [block, draw=warmdark,  fill=warmdark!80,inner xsep=2 * \bx, inner ysep=2 * \by]
\tikzstyle{flow}         = [->, >=latex, rounded corners, thick]

\foreach \l/\x in {1/\levOne, 2/\levTwo, 3/\levThree, 4/\levFour} {
    \node[encdec] (E_\l) at (-\x * \xsep + \xsep, \l * \ysep) {};
    \node[basicblock, anchor=south]   (EB_\l) at ([yshift=.55 * \by]E_\l) {$B^{\textrm{e}}_{\l}$};
    \node[downsampling, anchor=north] (ED_\l) at ([yshift=-.55 * \by]E_\l) {$D_{\l}$};
    \draw[->, >=latex] (EB_\l) -- (ED_\l);
}
\foreach \t/\f in {1/2, 2/3, 3/4} {
    \draw[flow] (ED_\f.south) |- (EB_\t.west);
}
\draw[flow,dashed,black!50] ([xshift=1.5*\bx]E_4.east) to[bend right=10] ([xshift=1.5*\bx,yshift=-\by]E_1.east);
\node[inner sep=0pt,align=center,text=black!50, fill=white] at ([xshift=2*\bx,yshift=-2*\by]E_3.east) {encoding};

\foreach \l/\x in {1/\levOne, 2/\levTwo, 3/\levThree, 4/\levFour} {
    \node[encdec] (D_\l) at (\x * \xsep - 1 * \xsep, \l * \ysep) {};
    \node[basicblock, anchor=south]   (DB_\l) at ([yshift=.55 * \by]D_\l) {$B^{\textrm{d}}_{\l}$};
    \node[upsampling, anchor=north] (DU_\l) at ([yshift=-.55 * \by]D_\l) {$U_{\l}$};
    \node[circle,draw,inner sep=0pt, fill=black!20] (C_\l) at ([xshift=-1.5*\bx]DB_\l.west) {$\oplus$};
    \draw[flow] (DU_\l) -| (C_\l);
    \draw[flow] (C_\l) -- (DB_\l);
}
\foreach \t/\f in {1/2, 2/3, 3/4} {
    \draw[flow] (DB_\t.east) -| (DU_\f.south) ;
}
\draw[flow,dashed,black!50] ([xshift=-3*\bx,yshift=-\by]D_1.west) to[bend right=10] ([xshift=-3*\bx]D_4.west);
\node[inner sep=0pt,align=center,text=black!50,fill=white] at ([xshift=-3*\bx,yshift=-2*\by]D_3.west) {decoding};

\draw[flow] (EB_4) --node[midway, fill=white] {\small $256,256,48$} (C_4);
\draw[flow] (EB_3) --node[midway, fill=white] {\small $128,128,96$} (C_3);
\draw[flow] (EB_2) --node[midway, fill=white] {\small $64,64,192$} (C_2);
\draw[flow] (EB_1) --node[midway, fill=white] {\small $32,32,384$} (C_1);

\node[bottleneck, text=white, inner sep=10] (N) at (0, 10pt) {pixel-wise ViT};
\draw[flow, rounded corners=10pt] (ED_1) |- node[xshift=-15pt, yshift=-7pt, pos=0.5] {\small $w=h=16,f=384$} (N);
\draw[flow, rounded corners=10pt] (N) -| node[right] {\small $16,16,384$} (DU_1);

\tikzstyle{processing} = [block,inner sep=5pt,draw=other,fill=other!80,align=center]
\node[processing, anchor=south] (Pre) at ([yshift=.5*\ysep]EB_4.north) {pre-process};
\node[inner sep=0,anchor=east] (I) at ([xshift=-1.5*\bx]Pre.west) {Input};
\node[inner sep=0,anchor=north west] at ([yshift=-\by]I.south west) {$256,256,3$};

\node[processing, anchor=south] (Post) at (Pre.south -| DB_4.north) {post-process};
\node[inner sep=0, anchor=west] (O) at ([xshift=1.5*\bx]Post.east) {Output};
\node[inner sep=0,anchor=north east] at ([yshift=-\by]O.south east) {$256,256,3$};
\draw[flow] (I) -- (Pre);
\draw[flow] (Pre) -- node[xshift=2pt, right,pos=.5] {$w_0=h_0=256,f_0=48$} (EB_4);
\draw[flow] (DB_4) --node[xshift=-2pt, left,pos=.5] {$256,256,48$} (Post);
\draw[flow] (Post) -- (O);

\node[block, draw=black!30, fill=black!10, anchor=south west,inner sep=3pt] (Leg) at ([xshift=-10pt]N.south -| D_3.east) {
\begin{tikzpicture}
    \node[basicblock, rounded corners=2, minimum width=.6 * \bx, minimum height=.6 * \by] (LBshape) at (0, 0) {$B$};
    \node[inner sep=0, anchor=west, font=\small] (LBtext) at ([xshift=5pt]LBshape.east) {basic block};
    \node[downsampling, rounded corners=2, minimum width=.6 * \bx, minimum height=.6 * \by, anchor=north] (LDshape) at ([yshift=-.8 * \by]LBshape.south) {$D$};
    \node[inner sep=0, anchor=north west, align=left, font=\small] (LDtext) at ([xshift=5pt]LDshape.north east) {down sampling: \\ \texttt{Conv}$(k=2,s=2)$};
    \node[upsampling, rounded corners=2, minimum width=.6 * \bx, minimum height=.6 * \by, anchor=north east] (LUshape) at ([yshift=-.8 * \by]LDshape.east |- LDtext.south) {$U$};
    \node[inner sep=0, anchor=north west, align=left, font=\small] (LUtext) at ([xshift=5pt]LUshape.north east) {
        up sampling:\\up-sample with\\scale $2$ and\\\texttt{Conv}$(k=3, p=1)$
    };
    \node[circle,draw,inner sep=0pt, fill=black!20, anchor=north] (LCshape) at ([yshift=-.8 * \by]LUshape |- LUtext.south) {$\oplus$};
    \node[inner sep=0, anchor=north west, align=left, font=\small] (LCtext) at (LCshape.north -| LUtext.west) {concatenation along\\feature dimension};
\end{tikzpicture}
};

\node[block, anchor=south east, draw=coollight!60, fill=coollight!10,inner sep=3pt] (B) at ([xshift=5pt]N.south -| EB_3.west) {
\begin{tikzpicture}
    \def\ys{-.8 * \by}
    \tikzstyle{basicblockdetail} = [block,anchor=north, font=\footnotesize, inner xsep=2.5pt, inner ysep=2.5pt, draw=coollight!60, fill=coollight!30]
    \node[basicblockdetail] (Norm) at (0, 0) {\texttt{InstanceNorm}};
    \node[basicblockdetail] (Conv) at ([yshift=\ys]Norm.south) {\texttt{Conv}$(k=3, p=1)$};
    \node[basicblockdetail] (Activ) at ([yshift=\ys]Conv.south) {\texttt{LeakyReLU}};
\end{tikzpicture}
};
\node[inner sep=0, anchor=south west] (BTitle) at ([yshift=.5 * \by]B.north west) {basic block};
\node[inner sep=0, anchor=south west] at ([xshift=2pt]B.south east) {$\times 2$};

\node[block, anchor=south, draw=other!60, fill=other!20,inner sep=3pt] (PreD) at ([yshift=1.5*\bx]B.north) {
\begin{tikzpicture}
    \def\ys{-.8 * \by}
    \tikzstyle{predetail} = [block,anchor=north, font=\footnotesize, inner xsep=2.5pt, inner ysep=2.5pt, draw=other, fill=other!80]
    \node[predetail] (Conv) at (0, 0) {\texttt{Conv}$(k=3, p=1)$};
    \node[predetail] (Activ) at ([yshift=\ys]Conv.south) {\texttt{LeakyReLU}};
\end{tikzpicture}
};
\node[inner sep=0, anchor=south west] at ([yshift=.5 * \by]PreD.north west) {pre-processing};

\node[block, anchor=south east, draw=other!60, fill=other!20,inner sep=3pt] (PostD) at ([yshift=1.5*\by]Leg.north east) {
\begin{tikzpicture}
    \def\ys{-.8 * \by}
    \tikzstyle{postdetail} = [block,anchor=north, font=\footnotesize, inner xsep=2.5pt, inner ysep=2.5pt, draw=other, fill=other!80]
    \node[postdetail] (Conv) at (0, 0) {\texttt{Conv}$(k=1)$};
    \node[postdetail] (Activ) at ([yshift=\ys]Conv.south) {\texttt{Sigmoid}};
\end{tikzpicture}
};
\node[inner sep=0, anchor=south west] at ([yshift=.5 * \by]PostD.north west) {post-processing};
\end{tikzpicture}


};
    \end{tikzpicture}}
    \caption{UNet ViT Generator with Full Details}
    \label{fig:unetvit_full}
\end{figure*}

\begin{figure*}
    \centering
    \resizebox{.95\textwidth}{!}{
    \begin{tikzpicture}
        \node[inner sep=0, fill=white] at (0, 0) {


\begin{tikzpicture}
\pgfdeclarelayer{back}
\pgfsetlayers{back,main}

\makeatletter
\pgfkeys{%
  /tikz/on layer/.code={
    \def\tikz@path@do@at@end{\endpgfonlayer\endgroup\tikz@path@do@at@end}%
    \pgfonlayer{#1}\begingroup%
  }%
}
\makeatother

\tikzstyle{block} = [
    rectangle, 
    rounded corners, 
    blur shadow={shadow blur steps=5},
    align=center,
    draw=black, 
	fill=black!5,
    inner sep=0pt,
    outer sep=0pt,
    rotate=90,
    anchor=north,
]

\def\bx{5pt}
\def\by{5pt}

\tikzstyle{transblock} = [block, draw=coollight!20, fill=coollight!20, inner xsep=1.25 * \bx, inner ysep=3 * \by]
\tikzstyle{unet}       = [block, draw=black!50, text=black!50, inner xsep=\bx,inner ysep=\by]
\tikzstyle{reshape}    = [block, draw=warmlight, fill=warmlight!80, inner xsep=\bx, inner ysep=\by]
\tikzstyle{linear}     = [block, draw=warmlight, fill=warmlight!80, inner xsep=\bx, inner ysep=\by]
\tikzstyle{norm}       = [block, text=white, draw=warmdark, fill=warmdark!80, inner xsep=\bx, inner ysep=\by]
\tikzstyle{attn}       = [block, text=white, draw=warmdark, fill=warmdark!80, inner xsep=\bx, inner ysep=\by]
\tikzstyle{ffn}        = [block, text=white, draw=warmdark, fill=warmdark!80, inner xsep=\bx, inner ysep=\by]
\tikzstyle{posemb}     = [block, draw=other, fill=other!80,, inner xsep=\bx, inner ysep=\by, rotate=-90]

\tikzstyle{plus}       = [circle,draw,inner sep=0pt,fill=black!20,anchor=north, rotate=90]
\tikzstyle{flow}       = [->, >=latex, rounded corners, thick]

\def\xs{26pt}
\def\sxs{.5 * \xs}
\node[reshape] (R1) at (0, 0) {reshape};
\node[plus] (C) at ([xshift=\xs]R1.south) {$\oplus$};
\node[posemb,anchor=north] (P) at ([yshift=-1.6 * \xs]C.east) {Positional\\Embedding};
\node[linear] (L1) at ([xshift=\xs]C.south) {\texttt{Linear}};
\node[norm] (TN1) at ([xshift=\xs]L1.south) {\texttt{LayerNorm}};
\node[attn] (A) at ([xshift=\sxs]TN1.south) {Multi-head self-\\attention (MSA)};
\node[plus] (Add1) at ([xshift=1.2 * \sxs]A.south) {$+$};
\node[norm] (TN2) at ([xshift=\sxs]Add1.south) {\texttt{LayerNorm}};
\node[ffn] (F) at ([xshift=.8 * \sxs]TN2.south) {Feed-Forward\\network (FFN)};
\node[plus] (Add2) at ([xshift=1.2 * \sxs]F.south) {$+$};
\node[linear] (L2) at ([xshift=.8 * \xs]Add2.south) {\texttt{Linear}};
\node[reshape] (R2) at ([xshift=\xs]L2.south) {reshape};

\tikzstyle{Fdetail} = [block, text=white, draw=warmdark, fill=warmdark!80, inner sep=2pt, rotate=-90, font=\fontsize{8}{8}\selectfont]
\node[Fdetail, draw=none, fill=warmdark!30,anchor=west,inner sep=4pt,minimum width=60pt] (Fd) at ([xshift=20pt,yshift=20pt]R2.south) {
\begin{tikzpicture}
    \def\ys{-.8 * \by}
    \node[Fdetail] (FdL1) at (0, 0) {\texttt{Linear}$(f_v, f_h)$};
    \node[Fdetail] (FdA) at ([yshift=\ys]FdL1.south) {\texttt{GeLu}};
    \node[Fdetail] (FdL2) at ([yshift=\ys]FdA.south) {\texttt{Linear}$(f_h, f_v)$};
    
\end{tikzpicture}
};
\node[inner sep=0,anchor=south west, font=\fontsize{8}{8}\selectfont] (FdTitle) at ([yshift=.5*\by]Fd.north west) {FFN};

\node[block, draw=black!30, fill=black!10, anchor=south east, rotate=-90, inner sep=2pt] (L) at (R2.south |- P.south) {
\begin{tikzpicture}
    \node[plus,rotate=-90] (LCshape) at (0, 0) {$\oplus$};
    \node[inner sep=0,anchor=west, align=left, font=\small] (LCtext) at ([xshift=.1 * \xs]LCshape.east) {Concatenation};
    \node[plus,rotate=-90,anchor=north] (LAshape) at ([yshift=-.1 * \xs]LCshape.south) {$+$};
    \node[inner sep=0,anchor=west, align=left, font=\small] (LAtext) at ([xshift=.1 * \xs]LAshape.east) {Addition};
\end{tikzpicture}
};

\tikzmath{
    coordinate \W;
    \W = (Fd.east) - (Fd.west); 
}
\tikzstyle{PEdetail} = [block,inner sep=2pt, draw=other, fill=other!80, rotate=-90, font=\fontsize{8}{8}\selectfont]
\node[PEdetail, anchor=south west, draw=other!60, fill=other!20, inner sep=4pt,minimum width=60pt] (PE) at (Fd.west |- L.south) {
\begin{tikzpicture}
    \def\ys{-.8 * \by}
    \node[PEdetail] (norm) at (0, 0) {normalization};
    \node[PEdetail] (linear) at ([yshift=\ys]norm.south) {\texttt{Linear}$(2, f_p)$};
    \node[PEdetail] (sine) at ([yshift=\ys]linear.south) {sine};
\end{tikzpicture}
};
\node[inner sep=0, anchor=south west,  font=\fontsize{8}{8}\selectfont] at ([yshift=.5 * \by]PE.north west) {positional embedding};

\tikzmath{
    coordinate \C;
    \C = (A.east -| Add2.south)-(TN1.north |- A.west); 
}
\begin{pgfonlayer}{back}
\node[block, draw=none, fill=warmdark!30,inner xsep=.66 * \Cy,inner ysep=.56 * \Cx,anchor=center] (transformerblock) at ($(TN1.north)!.46!(Add2.south)$) {};
\end{pgfonlayer}
\node[inner sep=2pt, anchor=north east] at ($(transformerblock.north east)!.98!(transformerblock.south east)$) {$\times12$};
\node[inner sep=0,anchor=south] at (P.south-| transformerblock.west) {Transformer encoder block};

\draw[flow] (R1) -- node[pos=.5,right,rotate=90,xshift=2pt] {\small$256, f$} (C);
\draw[flow] (C) -- node[pos=.5,right,rotate=90,xshift=2pt] {\small$256,f + f_p$} (L1);
\draw[flow] (P) -- node[pos=.55,above,rotate=90,yshift=4pt] {\small$256,f_p$} (C);
\draw[flow] (L1) -- node[pos=.6,right,rotate=90,xshift=2pt] {\small$256,f_v$} (TN1);
\draw[flow] (TN1) -- (A);
\draw[flow] (A) -- node[left,rotate=90,xshift=-2pt] {$\times\alpha$} (Add1);
\draw[flow] (Add1) -- (TN2);
\draw[flow] (TN2) -- (F);
\draw[flow] (F) -- node[left,rotate=90,xshift=-2pt] {$\times\alpha$} (Add2);
\draw[flow] (Add2) -- node[pos=.6,right,rotate=90,xshift=2pt] {\small$256,{f_v}$} (L2);
\draw[flow] (L2) -- node[pos=.5,right,rotate=90,xshift=2pt] {\small$256,f$} (R2);

\node[inner sep=0,outer sep=0] (ph1) at ([yshift=1pt]$(L1)!.5!(TN1)$) {};
\node[inner sep=0,outer sep=0] (ph2) at ([yshift=1pt]$(Add1)!.35!(TN2)$) {};
\draw[flow] (ph1) -- +(0,-1.8*\xs) -| (Add1.west);
\draw[flow] (ph2) -- +(0,-1.8*\xs) -| (Add2.west);

\end{tikzpicture}

};
    \end{tikzpicture}}
    \caption{Vision Transformer with Full Details}
    \label{fig:vit_full}
\end{figure*}
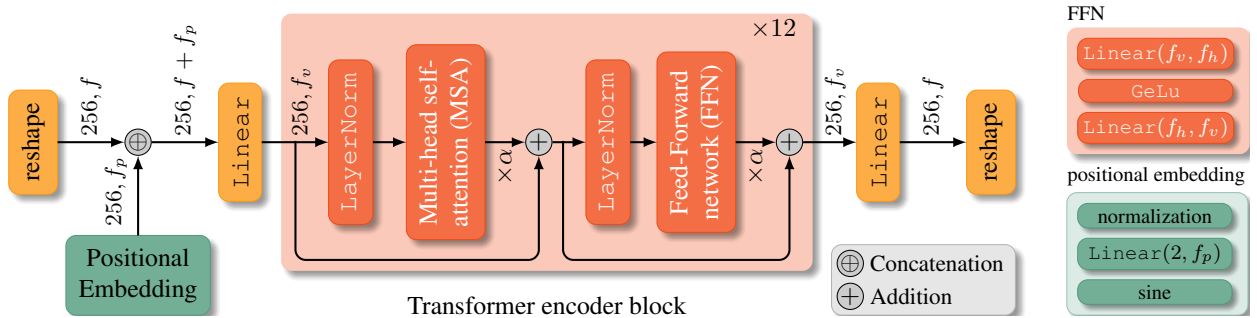

A \uv generator consists of a UNet~\cite{ronneberger2015u} with
a pixel-wise Vision Transformer (ViT)~\cite{dosovitskiy2020image} at the bottleneck (Figure~\ref{fig:unetvit_full}). 
UNet's encoding path extracts features from the input via four layers of convolution and downsampling. 
The features extracted at each layer are also passed to the corresponding layers of the decoding path via the skip connections, whereas the bottom-most features are passed to the pixel-wise ViT (Figure~\ref{fig:vit_full}).

On UNet's encoding path, the pre-processing layer turns an image into a tensor with dimension $(w_0,h_0,f_0)$. 
Each layer of the encoding path consists of a basic and downsampling block. 
The basic block is composed primarily of two convolutions, while the downsampling block has one convolution with stride $2$.
A pre-processed tensor will have its width and height halved at each downsampling block, while the feature dimension doubles at the last three downsampling blocks.  
Hence, the output from the encoding path will have dimension $(w,h,f) = (w_0/16, h_0/16, 8f_0)$, and it forms the input to the pixel-wise ViT bottleneck. 
Each layer of the UNet decoding path consists of an upsampling block followed by a basic block. 
A basic block on the decoding path differs from one on the encoding path in that it takes as input a concatenated tensor as input formed with the output from the upsampling layer and the tensor from the corresponding skip connection of the encoding path. 
The decoding path's output will go through a post-processing layer of $1\times1$ convolution with a sigmoid activation to produce an image.

A pixel-wise ViT is composed primarily of a stack of Transformer encoder blocks~\cite{devlin2019bert}.
To construct an input to the stack, the ViT first flattens an encoding along the spatial dimensions to form a sequence of transformer tokens. 
The token sequence has length $w \times h$, and each token in the sequence is a vector of length $f$. 
It then concatenates each token with its two-dimensional Fourier positional embedding~\cite{anokhin2021image} of dimension $f_p$ and linearly maps the result to have dimension $f_v$. 
To improve the Transformer convergence, we adopt the rezero regularization~\cite{bachlechner2021rezero} scheme and introduce a trainable scaling parameter $\alpha$ that modulates the magnitudes of the nontrivial branches of the residual blocks. 
The Transformer stack output is linearly projected back to have dimension $f$ and unflattened to have width $w$ and $h$.
In this study, we use input raw or cropped images with $w_0 = h_0 = 256$ and set $f_0 = 48$. 
Hence, we have $w = h = 16$ and $f = 384$. 
We use $12$ Transform encoder blocks in ViT and set $f_p, f_v = f$, and $f_h = 4f_v$ for the feed-forward network in each transformer encoder block.
\section{Additional Sample Translations}
\begin{figure*}[ht]
    \centering
    \includegraphics[width=.7\textwidth]{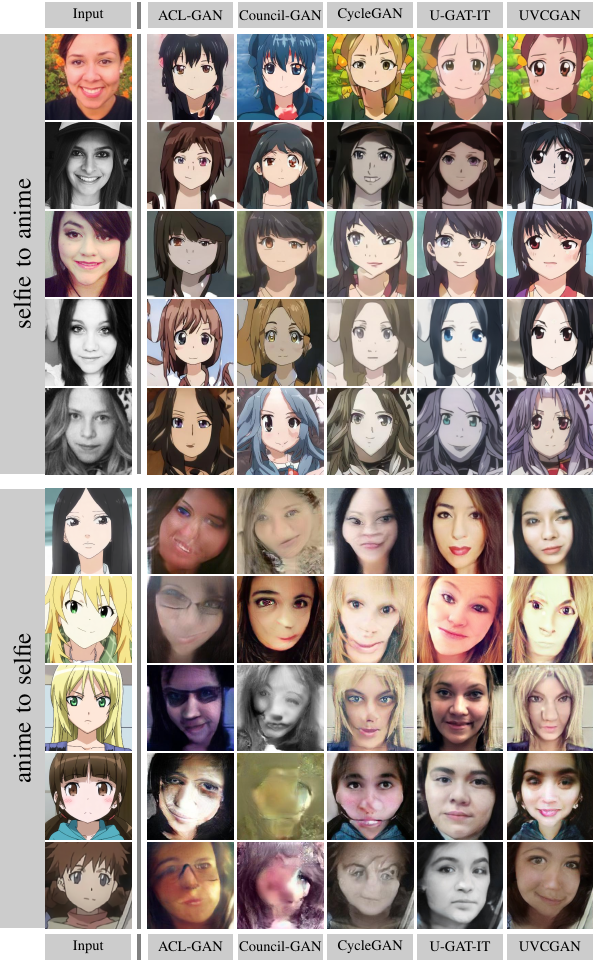}
    \caption{\textbf{Additional Sample Translations: \anime}}
    \label{fig:sup_image_grid_anime}
\end{figure*}

\begin{figure*}[ht]
    \centering
    \includegraphics[width=.7\textwidth]{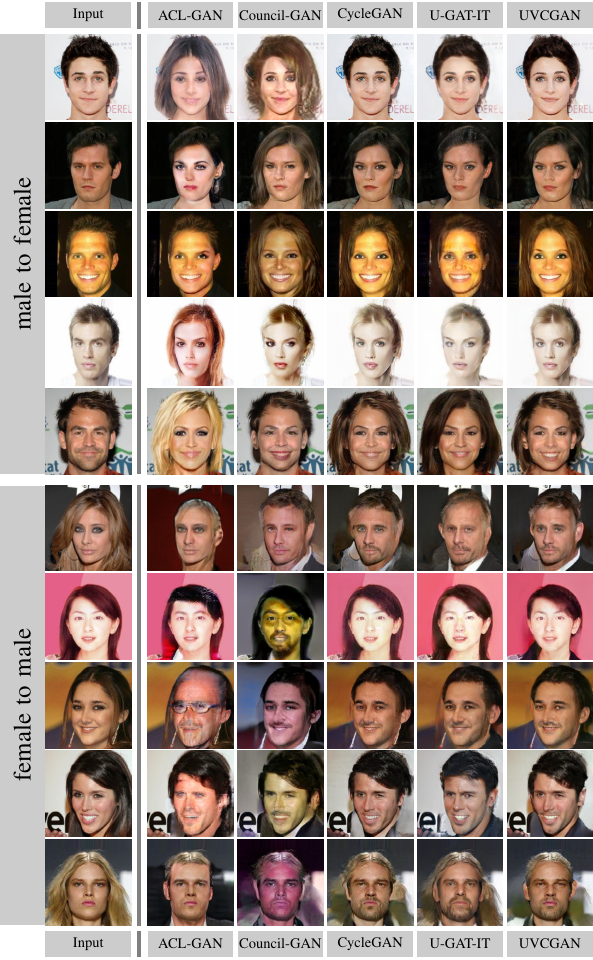}
    \caption{\textbf{Additional Sample Translations: \gender}}
    \label{fig:sup_image_grid_gender}
\end{figure*}

\begin{figure*}[ht]
    \centering
    \includegraphics[width=.7\textwidth]{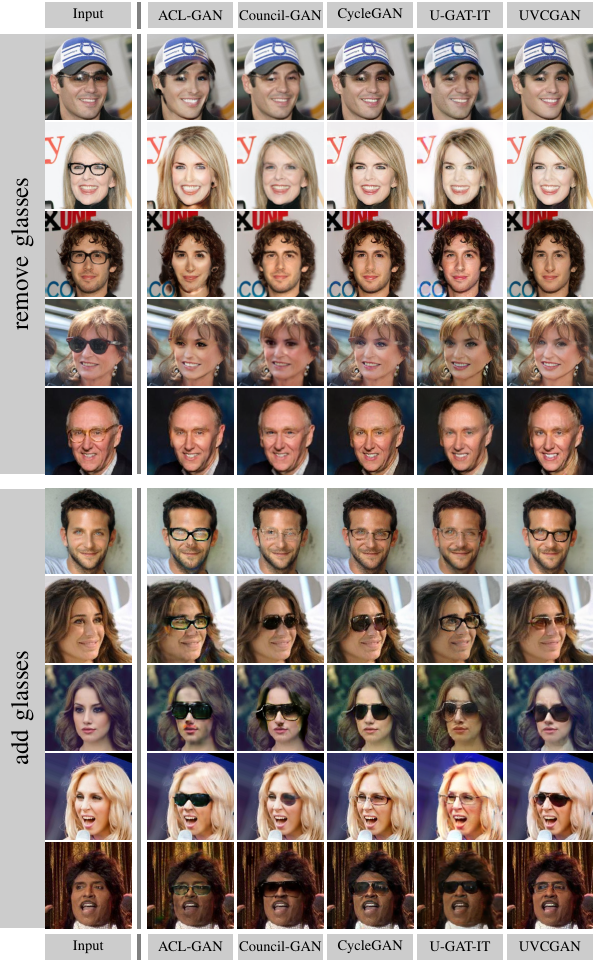}
    \caption{\textbf{Additional Sample Translations: \glasses}}
    \label{fig:sup_image_grid_glasses}
\end{figure*}

We show a few more translations in Figures~\ref{fig:sup_image_grid_anime} to \ref{fig:sup_image_grid_glasses}.

\end{appendices}

\end{document}